%% file: iclr2024_conference.tex
\documentclass{article} 
\usepackage{iclr2024_conference,times}

\input{math_commands.tex}

\usepackage{hyperref}
\usepackage{booktabs} 
\usepackage{url}
\usepackage{algorithm,algorithmic}
\usepackage{graphicx}
\usepackage{subfig}
\usepackage{xcolor}
\usepackage{wrapfig}
\usepackage{sidecap}
\usepackage{xspace}

\usepackage{colortbl}
\definecolor{Gray}{gray}{0.9}

\newcommand{\bl}[1]{\textcolor{black}{#1}}

\newcommand{\eg}{e.g.}

\usepackage[algo2e,ruled,vlined]{algorithm2e}

\usepackage{listings}
\title{Formatting Instructions for ICLR 2024 \\ Conference Submissions}
\title{Ask LLM More: Detecting LLM generated Content via Repharsing Difference}
\title{Prompting LLM for Machine Generated Text Detecting}
\title{Machine Generated Text Detection via }
\title{Detecting GPT by polishing it}
\title{Detecting  generated text by polishing it}
\title{Detecting  generated text by Improving it}
\title{Detecting  Generated Text by Rewriting it}
\title{Detecting  Generated Text via Rewriting}
\title{Raidar: geneRative AI Detection viA Rewriting}

\author{Chengzhi Mao$^1$ \& Carl Vondrick$^1$ \& Hao Wang$^2$ \& Junfeng Yang$^1$ \\
Columbia University$^1$ \ \ Rutgers University$^2$\\
}

\newcommand{\raidar}{Raidar\xspace}

\iclrfinalcopy 
\begin{document}

\maketitle

\begin{abstract}
We find that large language models (LLMs) are more likely to modify human-written text than AI-generated text when tasked with rewriting. This tendency arises because LLMs often perceive AI-generated text as high-quality, leading to fewer modifications. We introduce a method to detect AI-generated content by prompting LLMs to rewrite text and calculating the editing distance of the output. We dubbed our gene\textbf{R}ative \textbf{AI} \textbf{D}etection vi\textbf{A} \textbf{R}ewriting method \textbf{\raidar}.  \raidar  significantly improves the F1 detection scores of existing AI content detection models -- both academic and commercial -- across various domains, including News, creative writing, student essays, code, Yelp reviews, and arXiv papers, with gains of up to 29 points. Operating solely on word symbols without high-dimensional features, our method is compatible with black box LLMs, and is inherently robust on new content. Our results illustrate the unique imprint of machine-generated text through the lens of the machines themselves.


\end{abstract}

\input{files/def}

\input{files/intro}
\input{files/relatedwork}

\input{files/method}

\input{files/experiment}

\input{files/conclusion}

\bibliography{iclr2024_conference}
\bibliographystyle{iclr2024_conference}

\appendix

\input{files/appendix}

\end{document}

%% file: math_commands.tex
\usepackage{amsmath,amsfonts,bm}

\def\eqref#1{equation~\ref{#1}}

\def\1{\bm{1}}

\DeclareMathAlphabet{\mathsfit}{\encodingdefault}{\sfdefault}{m}{sl}
\SetMathAlphabet{\mathsfit}{bold}{\encodingdefault}{\sfdefault}{bx}{n}

\newcommand{\E}{\mathbb{E}}

\newcommand{\R}{\mathbb{R}}

\DeclareMathOperator*{\argmax}{arg\,max}
\DeclareMathOperator*{\argmin}{arg\,min}

%% file: files/def.tex
\def\Blue{\color{blue}}
\def\Purple{\color{purple}}

\def\A{{\bf A}}
\def\a{{\bf a}}
\def\B{{\bf B}}
\def\b{{\bf b}}
\def\C{{\bf C}}
\def\c{{\bf c}}
\def\D{{\bf D}}
\def\d{{\bf d}}
\def\E{{\bf E}}
\def\e{{\bf e}}
\def\f{{\bf f}}
\def\F{{\bf F}}
\def\K{{\bf K}}
\def\k{{\bf k}}
\def\L{{\bf L}}
\def\H{{\bf H}}
\def\h{{\bf h}}
\def\G{{\bf G}}
\def\g{{\bf g}}
\def\I{{\bf I}}
\def\R{{\bf R}}
\def\X{{\bf X}}
\def\Y{{\bf Y}}
\def\OO{{\bf O}}
\def\oo{{\bf o}}
\def\P{{\bf P}}
\def\Q{{\bf Q}}
\def\r{{\bf r}}
\def\s{{\bf s}}
\def\S{{\bf S}}
\def\t{{\bf t}}
\def\T{{\bf T}}
\def\x{{\bf x}}
\def\y{{\bf y}}
\def\z{{\bf z}}
\def\Z{{\bf Z}}
\def\M{{\bf M}}
\def\m{{\bf m}}
\def\n{{\bf n}}
\def\U{{\bf U}}
\def\u{{\bf u}}
\def\V{{\bf V}}
\def\v{{\bf v}}
\def\W{{\bf W}}
\def\w{{\bf w}}
\def\0{{\bf 0}}
\def\1{{\bf 1}}
\def\N{{\bf N}}

\def\AM{{\mathcal A}}
\def\EM{{\mathcal E}}
\def\FM{{\mathcal F}}
\def\TM{{\mathcal T}}
\def\UM{{\mathcal U}}
\def\XM{{\mathcal X}}
\def\YM{{\mathcal Y}}
\def\NM{{\mathcal N}}
\def\OM{{\mathcal O}}
\def\IM{{\mathcal I}}
\def\GM{{\mathcal G}}
\def\PM{{\mathcal P}}
\def\LM{{\mathcal L}}
\def\MM{{\mathcal M}}
\def\DM{{\mathcal D}}
\def\SM{{\mathcal S}}
\def\RB{{\mathbb R}}
\def\EB{{\mathbb E}}

\def\tx{\tilde{\bf x}}
\def\ty{\tilde{\bf y}}
\def\tz{\tilde{\bf z}}
\def\hd{\hat{d}}
\def\HD{\hat{\bf D}}
\def\hx{\hat{\bf x}}
\def\hR{\hat{R}}

\def\Ome{\mbox{\boldmath$\omega$\unboldmath}}
\def\bet{\mbox{\boldmath$\beta$\unboldmath}}
\def\et{\mbox{\boldmath$\eta$\unboldmath}}
\def\ep{\mbox{\boldmath$\epsilon$\unboldmath}}
\def\ph{\mbox{\boldmath$\phi$\unboldmath}}
\def\Pii{\mbox{\boldmath$\Pi$\unboldmath}}
\def\pii{\mbox{\boldmath$\pi$\unboldmath}}
\def\Ph{\mbox{\boldmath$\Phi$\unboldmath}}
\def\Ps{\mbox{\boldmath$\Psi$\unboldmath}}
\def\pss{\mbox{\boldmath$\psi$\unboldmath}}
\def\tha{\mbox{\boldmath$\theta$\unboldmath}}
\def\Tha{\mbox{\boldmath$\Theta$\unboldmath}}
\def\muu{\mbox{\boldmath$\mu$\unboldmath}}
\def\Si{\mbox{\boldmath$\Sigma$\unboldmath}}
\def\Gam{\mbox{\boldmath$\Gamma$\unboldmath}}
\def\gamm{\mbox{\boldmath$\gamma$\unboldmath}}
\def\Lam{\mbox{\boldmath$\Lambda$\unboldmath}}
\def\De{\mbox{\boldmath$\Delta$\unboldmath}}
\def\vps{\mbox{\boldmath$\varepsilon$\unboldmath}}
\def\Up{\mbox{\boldmath$\Upsilon$\unboldmath}}
\def\Lap{\mbox{\boldmath$\LM$\unboldmath}}
\newcommand{\ti}[1]{\tilde{#1}}

\def\tr{\mathrm{tr}}
\def\etr{\mathrm{etr}}
\def\etal{{\em et al.\/}\,}
\newcommand{\indep}{{\;\bot\!\!\!\!\!\!\bot\;}}
\def\argmax{\mathop{\rm argmax}}
\def\argmin{\mathop{\rm argmin}}
\def\vec{\text{vec}}
\def\cov{\text{cov}}
\def\dg{\text{diag}}

\newcommand{\tabref}[1]{Table~\ref{#1}}
\newcommand{\lemref}[1]{Lemma~\ref{#1}}
\newcommand{\thmref}[1]{Theorem~\ref{#1}}
\newcommand{\clmref}[1]{Claim~\ref{#1}}
\newcommand{\crlref}[1]{Corollary~\ref{#1}}
\newcommand{\eqnref}[1]{Eqn.~\ref{#1}}

\newtheorem{remark}{Remark}
\newtheorem{theorem}{Theorem}
\newtheorem{lemma}{Lemma}
\newtheorem{definition}{Definition}

\newtheorem{proposition}{Proposition}

%% file: files/intro.tex
\section{Introduction}


Large language models (LLMs) demonstrate exceptional capabilities in text generation~\citep{ChatGPT, GPT3, chowdhery2022palm}, such as question answering and executable code generation. The increasing deployment and accessibility of those LLM also pose serious risks~\citep{bergman2022guiding, mirsky2022threat}. For example, LLMs create cybersecurity threats, such as facilitating phishing attacks~\citep{kang2023exploiting},  generating propaganda~\citep{pan2023risk}, disseminating fake or biased content on social media, and lowering the bar for social engineering~\citep{asfour2023harnessing}. In education, they can lead to academic dishonesty~\citep{cotton2023chatting}. \citet{pearce2022asleep, siddiq2022empirical} have revealed that LLM-generated code can introduce security vulnerabilities to program. \citet{radford2023robust, shumailov2023curse} also find LLM-generated content is inferior to human content and can contaminate foundation models' training. Detecting and auditing those machine-generated text will thus be crucial to mitigate the potential downside of LLMs.

A plethora of works have investigated detecting machine-generated content~\citep{sadasivan2023can}. Early methods, including \cite{bakhtin2019real, fagni2021tweepfake, gehrmann2019gltr, ippolito2019automatic, jawahar2020automatic}, were effective before the emergence of sophisticated GPT models, yet the recent LLMs have made traditional heuristic-based detection methods increasingly inadequate~\cite{verma2023ghostbuster, gehrmann2019gltr}. Current techniques~\citep{mitchell2023detectgpt, verma2023ghostbuster} rely on LLM's numerical output metrics. \cite{gehrmann2019gltr, ippolito2019automatic, solaiman2019release} use token log probability. However, those features are not available in black box models, including state-of-the-art ones (\eg, GPT-3.5 and GPT-4). Furthermore, the high-dimensional features employed by existing methods often include redundant and spurious attributes, leading the model to overfit to incorrect features.

In this paper, we present \raidar,  a simple and effective method for detecting machine-generated text by prompting LLMs to rewrite it. Similar to how humans prompt LLMs for coherent and high-quality text generation, our method uses rewriting prompts to gain additional contextual information about the input for more accurate detection. 



\begin{figure}[t]
\centering
\vspace{-5mm}
\includegraphics[width=\textwidth]{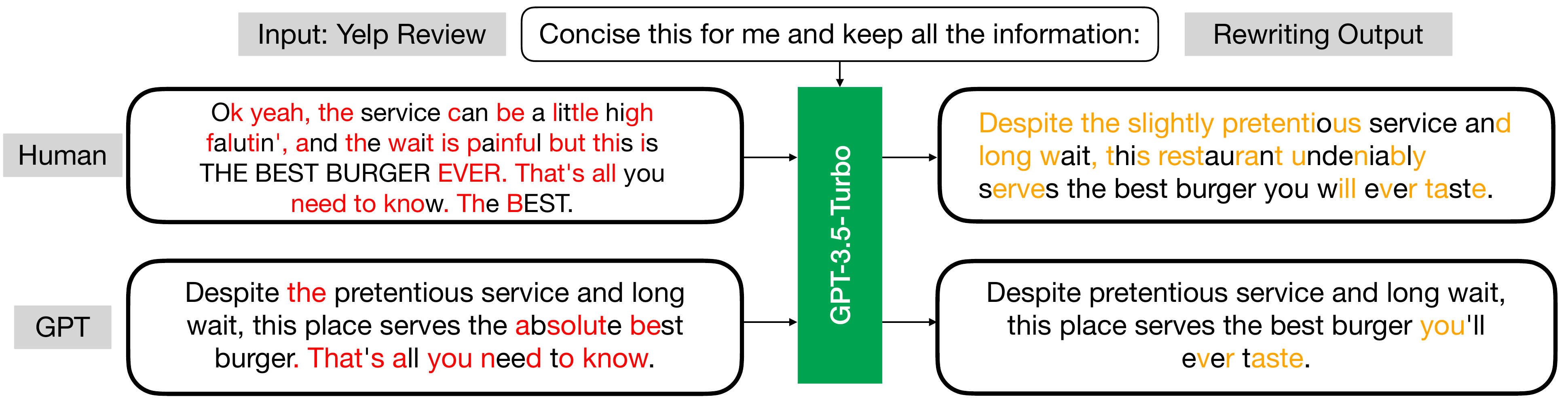}

  \caption{We introduce ``Detecting via Rewriting,'' an approach that detects machine-generated text by calculating rewriting modifications. We show the character deletion in \textcolor{red}{red} and the character insertion in \textcolor{orange}{orange}. Human-generated text tends to trigger more modifications than machine-generated text when asked to be rewritten. Our method is simple and effective, requiring the least access to LLM while being robust to novel text input. }
  \label{fig:teaser}
  \vspace{-5mm}
\end{figure}

Our key hypothesis is that text from auto-regressive generative models retains a consistent structure, which another such model will likely to also have a low loss and treat it as high quality. We observe that machine-generated text is less frequently altered upon rewriting compared to human-written text, regardless of the models used; see Figure~\ref{fig:teaser} as an example. Our approach \raidar shows how to capitalize on this insight to create detectors for machine-generated text. \raidar operates on the symbolic word output from LLMs, eliminating the need for deep neural network features, which boosts its robustness, generalizability, and adaptability. By focusing on the character editing distance between the original and rewritten text, \raidar is semantically agnostic, reducing irrelevant and spurious correlations. This feature-agnostic design also allows for seamless integration with the latest LLM models that only provide word output via API.  Importantly, our detector does not require the original generating model, allowing model A to detect the output of model B.



Visualizations, empirical experiments show that our simple rewriting-based algorithm \raidar significantly improves detection for several established paragraph-level detection benchmarks. \raidar advances the state-of-the-art detection methods~\citep{verma2023ghostbuster, mitchell2023detectgpt} by up to 29 points. Our method generalizes to six different datasets and domains, and it is robust when detecting text generated from different language models, such as Ada, Text-Davinci-002, Claude, and GPT-3.5, even though the model has never been trained on text generated from those models. In addition, our detection remains robust even when the text generation is aware of our detection mechanism and uses tailored prompts to bypass our detection. Our data and code is available at \url{https://github.com/cvlab-columbia/RaidarLLMDetect.git}.



%% file: files/relatedwork.tex
\section{Related Work}
\vspace{-3mm}
\textbf{Machine Text Generation.} Machine generated text has achieved high quality as model improves~\citep{radford2019language, li2022pretrained, zhou2023recurrentgpt, zhang2022opt, gehrmann2019gltr, GPT3, chowdhery2022palm}. The release of ChatGPT enables instructional following text synthesis for the public~\cite{ChatGPT}. \citep{dou, jawahar2020automatic} demonstrate that machines can potentially leave distinctive signals in the generated text, but these signals can be difficult to detect and may require specialized techniques.

\textbf{Detecting Machine Generated Text.} 
Detecting AI-generated text has been studied before the emergence of LLM~\citep{bakhtin2019real, fagni2021tweepfake, gehrmann2019gltr, ippolito2019automatic}. \citet{jawahar2020automatic} provided a detailed survey for machine-generated text detection.  

The high quality of recent LLM generation makes detection to be challenging~\citep{verma2023ghostbuster}. \citet{chakraborty2023possibilities} studies when it is possible to detect LLM-generated content. \citet{tang2023science} surveys literature for detecting LLM generated texts. \citet{sadasivan2023can} show that the detection AUROC is upper bounded by the gap between the machine text and human text. The state-of-the-art LLM detection algorithm~\citep{verma2023ghostbuster, mitchell2023detectgpt} requires access to the probability and loss output from the LLM \bl{for the scoring model}, yet those numerical metrics and features are not available for the latent GPT-3.5 and GPT-4. \bl{\citet{mitchell2023detectgpt} requires the scoring model and the target model to be the same. Ghostbuster~\citep{verma2023ghostbuster} operates under the assumption that the scoring and target model are different, but it still requires access to generated documents from the target model.} In addition, the output from the above deep scoring models can contain nuisances and spurious features, and can also be manipulated by adversarial attacks~\citep{jin2019bert, zou2023universal}, making detection not robust. Another line of work aims to watermark the AI-generated text to enable detection~\citep{kirchenbauer2023watermark}.


\textbf{Bypassing Machine Text Detection.} \citet{krishna2023paraphrasing} showed rephrase can remove watermark.
\citet{krishna2023paraphrasing, sadasivan2023can} show that paraharase can efficiently evade detection, including DetectGPT~\citep{mitchell2023detectgpt}, GLTR~\citep{gehrmann2019gltr}, OpenAI's generated text detectors, and other zero-shot methods~\cite{ippolito2019automatic, solaiman2019release}. There is a line of work that watermarks the generated text to enable future detection. However, they are shown to be easily broken by rephrasing, too. Our detection can be robust to rephrasing.

\textbf{Prompt Engineering.} Prompting is the most effective and popular strategy to adapt and instruct LLM to perform tasks~\citet{li2021prefix, zhou2022large, wei2022chain, kojima2022large}. Zero-shot GPT prompts the GPT model by asking ``is the input generated by GPT'' to predict if this is GPT generated~\citep{verma2023ghostbuster}. However, since GPTs are not trained to perform this task, they struggle. In contrast, our work constructs a few rewriting prompts to access the inherent invariance and equivariance of the input. While we can also perform an optimization-based search for better prompt~\citep{zhou2022large}, we leave this for future work.

%% file: files/method.tex
\section{Detecting Machine Generated Text by Rewriting}

\begin{figure*}[t]
\vspace{-9mm}
\centering
\subfloat[Invariance]{\label{aasxs}\includegraphics[width=0.33\textwidth]{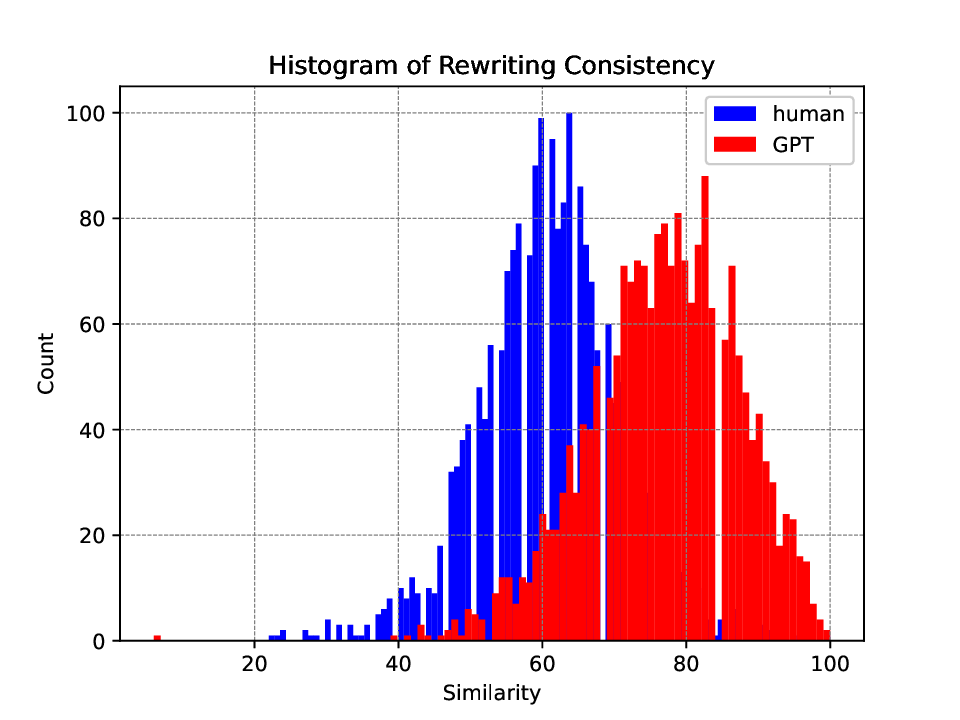}}
\subfloat[Equivariance]{\label{aasxs}\includegraphics[width=0.33\textwidth]{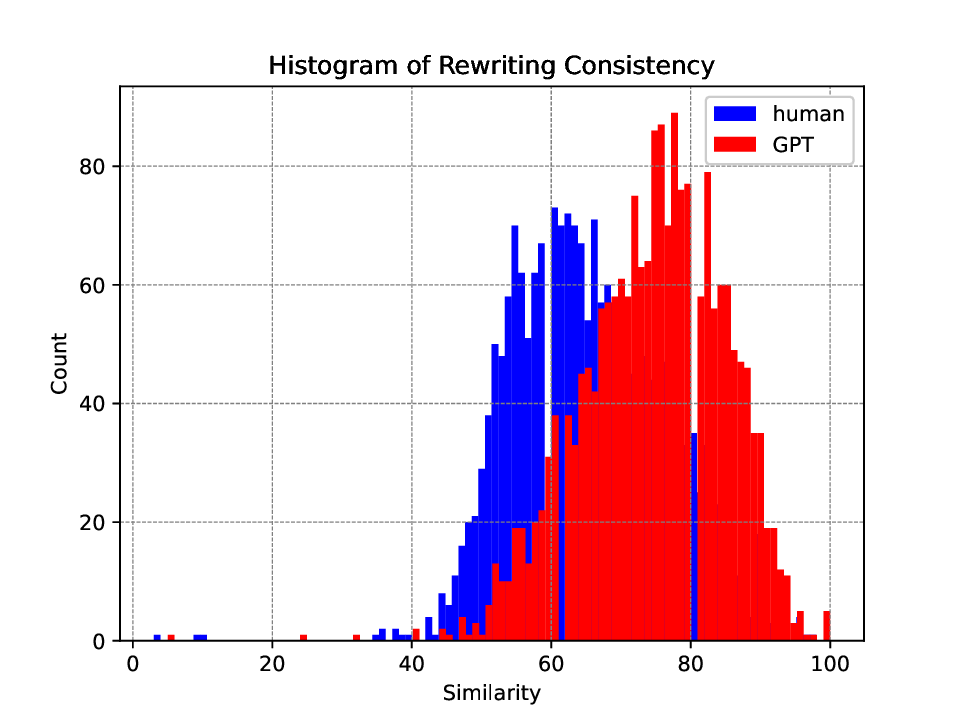}}
\subfloat[Uncertainty]{\label{aasxs}\includegraphics[width=0.33\textwidth]{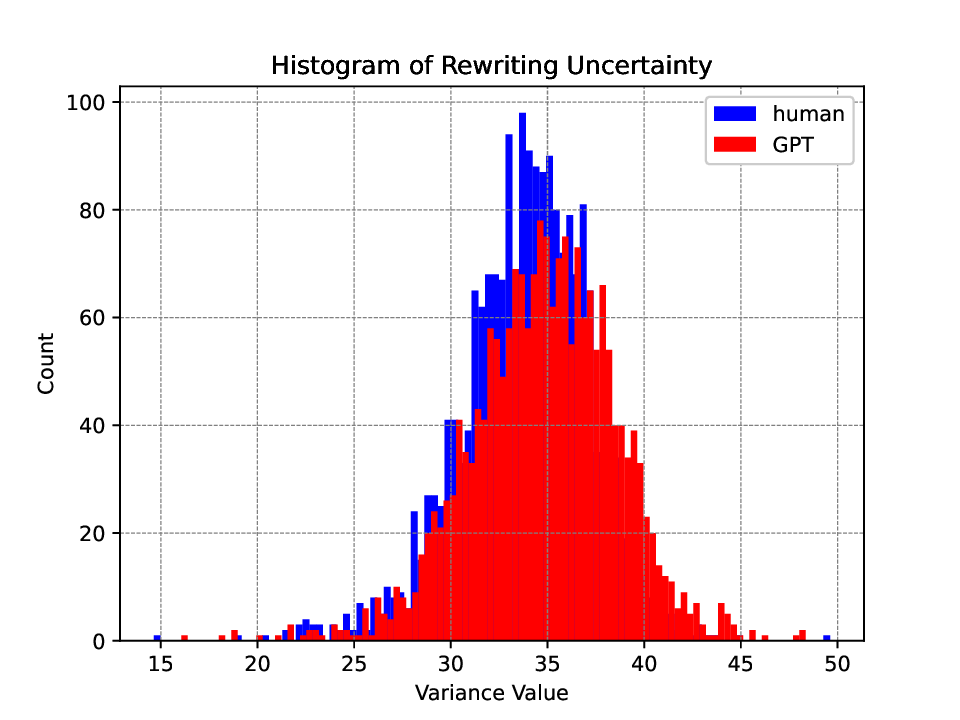}}
\caption{The rewriting similarity score of human and GPT-generated text. The similarity score measures how similar the text is before and after the rewriting. A larger similarity score indicates that rewriting makes less change. (a) We show the similarity score under a single transformation; machine-generated text (red) is invariant after rewriting compared with human-generated text. (b) We show the similarity score under a transformation and its reverse transformation; the machine-generated text is more equivariant under transformation. (c) We show the uncertainty of text produced by humans and GPT. GPT input is more stable than human input. The samples are run on the Yelp Review dataset with 4000 samples. The discrepancies in invariance, equivariance, and output uncertainty allow us to detect machine-generated text.} 
\label{fig:curve}
\vspace{-3mm}
\end{figure*}

We present our approach \raidar for detecting large language models generated text via rewriting. We first talk about the rewriting prompt design to access the property of the input text, then introduce our approach that detects based on the output symbolic modifications.

\subsection{Rewriting Text via Language Models and Prompts}

Let $F(\cdot)$ be a large language model. Given an input text $\x$, our goal is to classify the label $\y$, which indicates whether it is generated by a machine. 
The key observation of our method is that given the same rewriting prompt, such as asking the LLM model to ``rewrite the input text,'' an LLM-written text will be accepted by the language model as a high-quality input with inherently lower loss, which leads to few modifications at rewriting. In contrast, a human-written text will be unfavoured by LLM and edited more by the language models.

We will use the invariance between the output and the input to measure how much LLM prefers the given input. We hypothesize that LLM will produce invariant output when rewriting its own generated text because another auto-regressive prediction will tend to produce text in a similar pattern. We define this property as the invariance property.



\begin{figure}[t]
\centering

\includegraphics[width=\textwidth]{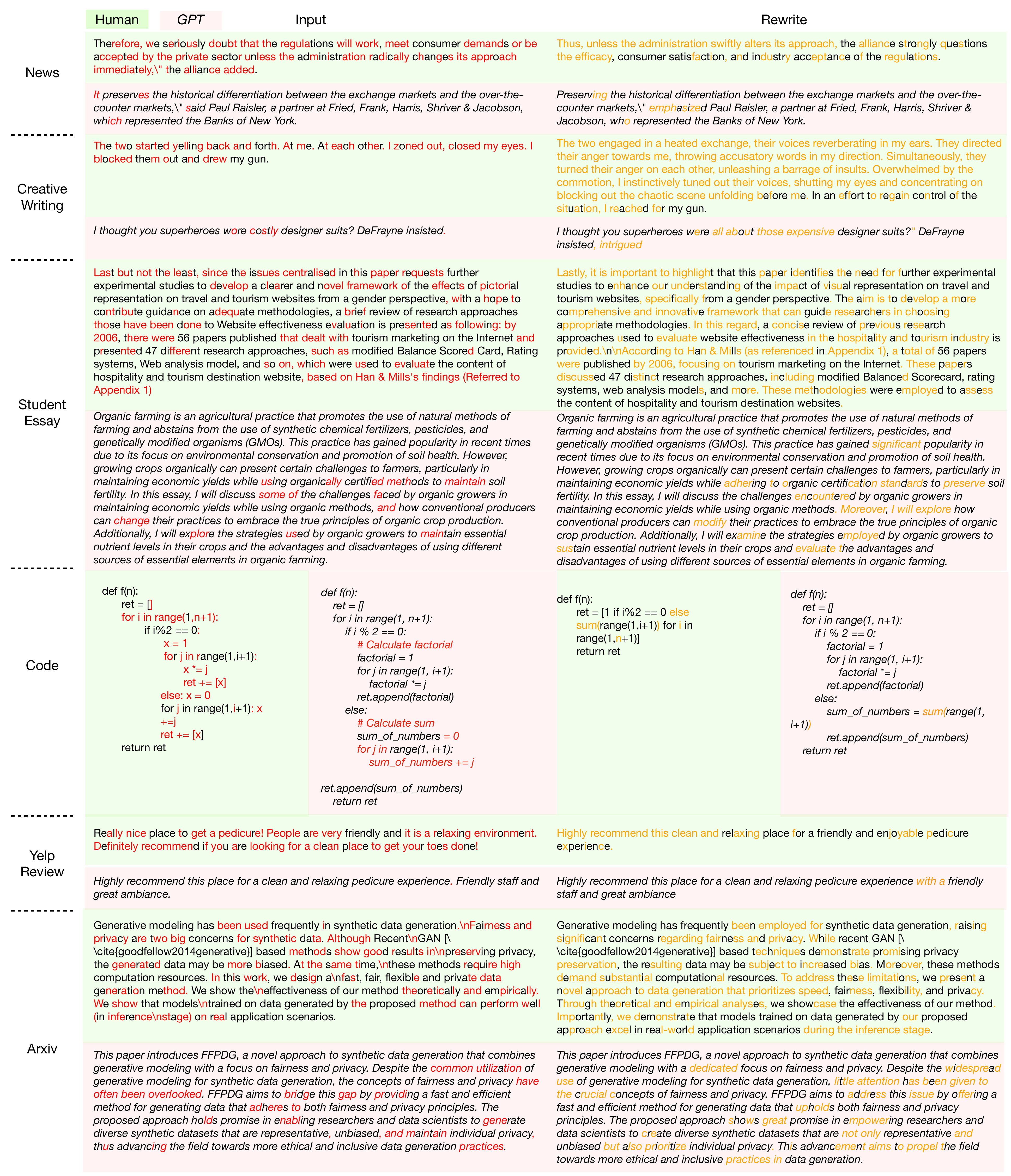}

  \caption{Examples of text rewriting on six datasets for invariance. We use a green background to indicate human-written text, and a red background to indicate machine-generated text. We show the character deletion in \textcolor{red}{red} and the character insertion in \textcolor{orange}{orange}. Human-written text tends to be modified more than machine-generated text. Our detection algorithm relies on this difference to make predictions.}
  \vspace{-5mm}
  \label{fig:}
\end{figure}

\textbf{Invaraince.} Given data $x$, we apply a transformation to the data via prompting the LLM with prompt $p$. If the data $x$ is produced from LLM, then the transformation $p$ that aims to rewrite the input should introduce a small change. We construct the invariance measurement as $L = D(F(p,\x), \x)$, where $D(\cdot)$ denotes the modification distance.

We manually create the prompt $p$ to access this invariance. We do not study automatic ways to generate prompts~\cite{zhou2022large, li2021prefix}, which can be done in future work by optimizing the prompt. In this work, we will show that even a single manually written prompt can achieve a significant difference in invariance behavior. We show a few of our prompts here:
\begin{lstlisting}[breakatwhitespace=true]
1. Help me polish this:
2. Rewrite this for me:
3. Refine this for me please:
\end{lstlisting}
where the goal is to make LLM modify more when rewriting human text and be more invariant when modifying LLM-generated text.

\textbf{Equivariance.} In addition, we hypothesize that GPT data will be equivariant to the data generated by itself. Equivariance means that, if we transform the input, perform the rewriting, and undo the transformation, it will produce the same output as directly rewriting the input.


We achieve the transformation for large language models by appending a prompt $T$ to the input and asking the LLM to produce the transformed output. We denote the reversal of the transformation as $T^{-1}$, which is another prompt that writes in the opposite way as $T$. Equivariance can be measured by the following distance: $ L = D(F(T^{-1}, F(p, F(T, \x))), F(p, \x))$.


Here we show two examples for the equivariance transformation prompt $T$ and $T^{-1}$:
\begin{lstlisting}[breakatwhitespace=true,basicstyle=\ttfamily,mathescape=true]
$T$: Write this in the opposite meaning: 
$T^{-1}$: Write this in the opposite meaning:

$T$: Rewrite to Expand this:
$T^{-1}$:  Rewrite to Concise this:
\end{lstlisting}
By rewriting the sentence with the opposite meaning twice, the sentence should be converted back to its original if the LLM is equivariant to the examples. Note that this transformation $T$ is based on the language model prompt. 


\begin{figure}[t]
\centering
\includegraphics[width=\textwidth]{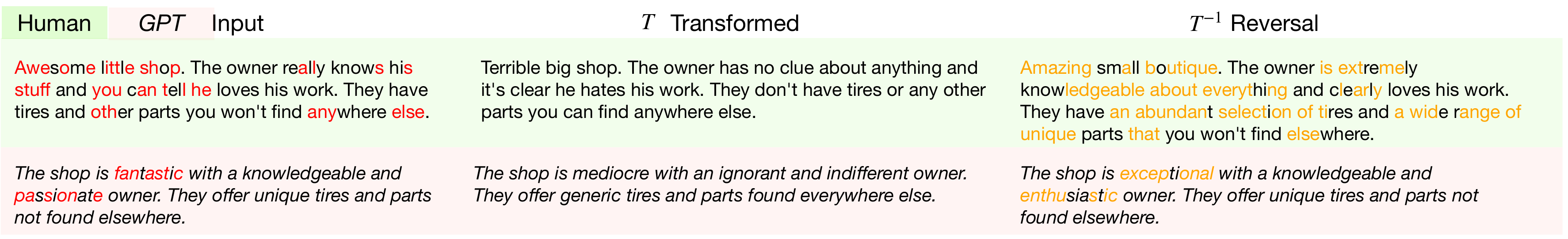}
  \caption{Examples for equivariance. We show an example on the Yelp Review dataset. For simplicity, we use identity transformation $p$, and use the ``opposite meaning'' as the equivariance transformation $T$. GPT data tends to be consistent to the original input after transformation and reversal.} 
  \vspace{-5mm}
  \label{fig:equi}
\end{figure} 

\textbf{Output Uncertainty Measurement.} We also assume that LLM-generated text will be more stable, when asked to rewrite multiple times than human-written text. We thus explore the variance of the output as a detection measurement.
Denote the prompt to be $p$. The k-th generation results from LLM would be $\x'_k = F(p, \x)$. Due to the randomness in language generation, $\x'_k$ will be different. We denote the editing distance between two outputs $A$ and $B$ as $D(A, B)$. We construct the uncertainty measurement as:$U = \sum_{i=1}^{K-1}\sum_{j=i}^{K} D(\x_i', \x_j')$. Note that, in contrast to the invariance and equivariance, this metric only uses the output, and the original input is not in the calculation of the output uncertainty.


\subsection{Measuring Change in Rewriting}

We treat the output of LLM as symbolic representations that encode information about the data.
In contrast to \citet{mitchell2023detectgpt, verma2023ghostbuster}, our detection algorithm does not use continuous, numerical representations of the word tokens. Instead, our algorithm operates totally on the discrete, symbolic representations from the LLM. By prompting LLM, our method obtains additional information about the input text via the rewriting difference. We will show how to measure the rewriting change below:

\textbf{Bag-of-words edit.} We use the change of bag-of-words to capture the edit created by LLM. We compute the number of common bags of n-words divided by the length of the input.

\textbf{Levenshtein Score}. Levenshtein score~\citep{levenshtein1966binary} is a popular metric for measuring the minimum number of single-character edits, including deletion and addition, to change one string to the other. We use standard dynamic programming to calculate the Levenshtein distance. A higher score denotes the two strings are more similar. We use $\text{Levenshtein}(A,B)$ to denote the edit distance between string $A$ and $B$. Let the rewriting output $\s_k = F(p_k, \x)$.
We obtain the ratio via:
\begin{equation*}
    D_k(\x, \s_k) = 1 - \frac{\text{Levenshtein}(\s_k, \x)}{\max(len(\s_k), len(\x)}.
\end{equation*}
We use ratio because the feature of editing difference should be independent of the text length. The invariance, equivariance, and uncertainty measured by the above metric will be used as features for a binary classifier, which predicts the generation source of the text. For details of the algorithm, please refer to Appendix~\ref{sec:alg}.



Our design enjoys several advantages. First, since we only access the discrete token output from LLM, our algorithm requires minimal access to the LLM models. Given that the major state-of-the-art LLM models, like GPT-3.5-turbo and GPT-4 from OpenAI, are black-box models and only provide API for accessing the discrete tokens rather than the probabilistic values, our algorithm is general and compatible with them. Second, since our representation is discrete, it is more robust in the sense that it will be invariant to the perturbations and shifting in the input space. Lastly, our symbolic representations enable us to construct the following measurements that are none differentiable, which introduces extra burden and cost for gradient-based adversarial attempts to bypass our detection model.


%% file: files/experiment.tex
\section{Results}
We conduct experiments on detecting AI-generated text on paragraph level and compare it to the state of the art. To further understand factors that affect detection performance, we also study the robustness of our method under input aiming to evade our detection, detection accuracy on text generated from different LLM sources, and evaluate our method with different LLM for rewriting.



\begin{table}[t]
\caption{F1 score for detecting machine-generated paragraphs. The results are in domain testing, where the model has been trained on the same domain. We \textbf{bold} the best performance on in-distribution and out-of-distribution detection. Our method achieved over 8 points of improvement over the established state-of-the-art.}
\vspace{-5mm}
\label{tab:main1}
\begin{center}
\begin{tabular}{l|cccccc}
\toprule
\multicolumn{1}{c|}{}  & \multicolumn{6}{c}{ Datasets} \\
\multicolumn{1}{c|}{} & & Creative & Student & & Yelp & Arxiv \\
\multicolumn{1}{c|}{Methods}  & News &  Writing &  Essay & Code &  Reviews &  Abstract   \\

\midrule
GPT Zero-Shot~\cite{verma2023ghostbuster} & 54.74 & 20.00 & 52.29& 62.28 & 66.34 & 65.94 \\
GPTZero~\citep{gptzero} & 49.65 & 61.81 & 36.70 & 31.57 & 25.00 & 45.16\\
DetectGPT~\cite{mitchell2023detectgpt} & 37.74 & 59.44 & 45.63 & 67.39 & 69.23 & 66.67\\
Ghostbuster~\cite{verma2023ghostbuster} & 52.01 & 41.13 & 42.44 & 65.97 & 71.47 & 76.82 \\
\cellcolor{Gray}Ours (Invariance) & \cellcolor{Gray}\textbf{60.29} & \cellcolor{Gray}\textbf{62.88} & \cellcolor{Gray}\textbf{64.81} & \cellcolor{Gray}\textbf{95.38} & \cellcolor{Gray}\textbf{87.75} & \cellcolor{Gray}81.94  \\
\cellcolor{Gray}Ours (Equivariance)  &  \cellcolor{Gray}58.00 & \cellcolor{Gray}60.27 &\cellcolor{Gray}60.07 &\cellcolor{Gray}80.55 &\cellcolor{Gray}83.50 &\cellcolor{Gray}75.74\\
\cellcolor{Gray}Ours (Uncertainty)  &\cellcolor{Gray}60.27 &\cellcolor{Gray}60.27 &\cellcolor{Gray}57.69 &\cellcolor{Gray}77.14 & \cellcolor{Gray}81.79 & \cellcolor{Gray}\textbf{83.33} \\
\bottomrule
\end{tabular}
\end{center}
\end{table}

\begin{table}[t]
\caption{F1 score for detecting machine-generated paragraph following the out-of-distribution setting in \citet{verma2023ghostbuster}. We use logistic regression classifier for all ours. Our method achieved over 22 points of improvement over the established state-of-the-art.}
\vspace{-5mm}
\label{tab:main_ood}
\begin{center}
\begin{tabular}{l|ccc}
\toprule
\multicolumn{1}{c|}{}  & \multicolumn{3}{c}{ Datasets} \\
\multicolumn{1}{c|}{Methods}  & News & Creative Writing & Student Essay  \\

\midrule

Ghostbuster~\cite{verma2023ghostbuster} & 34.01 & 49.53 & 51.21 \\
\cellcolor{Gray}Ours (Invariance)  & \cellcolor{Gray}56.47 & \cellcolor{Gray}55.51 & \cellcolor{Gray}\textbf{52.77}  \\
\cellcolor{Gray}Ours (Equivariance)  & \cellcolor{Gray}\textbf{56.87} & \cellcolor{Gray}\textbf{59.47} & \cellcolor{Gray}51.34  \\
\cellcolor{Gray}Ours (Uncertainty) &  \cellcolor{Gray}55.04 & \cellcolor{Gray}52.01 & \cellcolor{Gray}47.47 \\
\bottomrule
\end{tabular}
\end{center}
\vspace{-5mm}
\end{table}


\subsection{Dataset}
To evaluate our approach to the challenging, paragraph-level machine-generated text detection, we experiment with the following datasets.

\textbf{Creative Writing Dataset} is a language dataset based on the subreddit WritingPrompts, which is creative writing by a community based on the prompts. We use the dataset generated by \citet{verma2023ghostbuster}. We focus on detecting paragraph-level data, which is generated by text-davinci-003. 

\textbf{News Dataset} is based on the Reuters 50-50 authorship identification dataset. We use the machine-generated text from \citet{verma2023ghostbuster} via text-davinci-003.

\textbf{Student Essay Dataset}
The dataset is based on the British Academic Written English corpus and generated by \citet{verma2023ghostbuster}.

\textbf{Code Dataset.} The goal is to detect if the Python code has been written by GPT, which can be important for education. We adopt the HumanEval dataset~\citep{chen2021codex} as the human-written code, and ask GPT-3.5-turbo to perform the same task and generate the code.

\textbf{Yelp Review Dataset.}  Yelp reviews tend to be short and challenging to detect. We \bl{use the first} 2000 human reviews from the Yelp Review Dataset, and generate concise reviews via GPT-3.5-turbo in a similar length as the human written one. 

\textbf{ArXiv Paper Abstract.} We investigate if we can detect GPT written paragraphs in academic papers. Our dataset contains 350 abstracts from ICLR papers from 2015 to 2021, which are human-written texts since ChatGPT was not released then. We use GPT-3.5-turbo to generate an abstract based on the paper's title and the first 15 words from the abstract.

\subsection{Baselines}
\textbf{GPT Zero-shot}~\citep{verma2023ghostbuster} performs detection by directly asking GPT if the input is written by GPT or not. We use the same prompt as \citet{verma2023ghostbuster} to query GPT.

\textbf{GPTZero}~\citep{gptzero} is an commercial machine text detection service.

\textbf{DetectGPT}~\citep{mitchell2023detectgpt} is the state-of-the-art thresholding approach to detect GPT-generated text, which achieved 99-point performance over a longer input context, yet its performance on shorter text is unknown. It thresholds the curvature of the input to perform detection. \bl{We use the facebook/opt-2.7B for the scoring model.}

\textbf{Ghostbuster}~\citep{verma2023ghostbuster} is the state-of-the-art classifier for machine generated text detection. It uses probabilistic output from large language models as features, and performs feature selection to train an optimal classifier.

\begin{table}[t]
\caption{Performance under adaptive prompts aiming to evade our detector. In the "Single Training Prompt" column, the detector is trained on a non-adaptive prompt and tested against both the same prompt and two evasive prompts. Adversarial rephrasing can bypass our detector. In "Multi Training Prompt*", the model is trained using two prompts and tested on a third, different prompt. The last two rows shows results under adaptive prompts to evade our detection. Training on multiple prompts enhances our detector's robustness against machine-generated inputs attempting evasion.}
\label{tab:adaptive_attack}
\begin{center}
\begin{tabular}{l|ccc|ccc}
\toprule
 & \multicolumn{3}{c|}{Single Training Prompt}  & \multicolumn{3}{c}{Multi Training Prompt*} \\
Test Prompt  & Code & Yelp& Arxiv & Code & Yelp & Arxiv  \\
\midrule
No Adaptive Prompt & 95.38 & 87.75 &  81.94 & 92.76 & 58.04 & 82.25 \\
Prompt 1 to bypass detection & 34.15 & 61.38 & 43.81 & 86.95 & 69.19 & 91.89\\
Prompt 2 to bypass detection & 25.64 & 61.38 & 50.90 & 88.88 & 73.23 & 93.06\\
\bottomrule
\end{tabular}
\end{center}
\vspace{-5mm}
\end{table}


\subsection{Main Results}

We use GPT-3.5-Turbo as the LLM to rewrite the input text. Once we obtain the editing distance feature from the rewriting, we use Logistic Regression~\citep{berkson1944application} or XGBoost~\citep{chen2016xgboost} to perform the binary classification.
We compare our results on three datasets from~\citet{verma2023ghostbuster}, as well as our created three datasets, in Table~\ref{tab:main1}.   Our method \raidar outperforms the Ghostbuster method by up to 29 points, which achieves the best results over all baselines.
In Table~\ref{tab:main_ood}, we follow the out-of-distribution (OOD) experiment setup in \citet{verma2023ghostbuster}, where we trained the detection classifier on one dataset and evaluated on the other. For the OOD experiment, our method still improves by up to 32 points, demonstrating the effectiveness of our approach over prior methods.







\subsection{Analysis}

\begin{table}[t]
\vspace{-5mm}
\caption{Robustness in detecting outputs from various language models. Using the same GPT-3.5-Turbo rewriting model, we present F1 detection scores for detecting text from \bl{five} generation models across three diverse tasks. In the in-distribution experiment, detectors are trained and tested on the same model. For out-of-distribution, detectors are trained on text from other generators. Overall, our method effectively detects machine-generated text in both scenarios.}
\vspace{-3mm}
\label{tab:diffsource}
\begin{center}
\begin{tabular}{l|ccc|ccc|ccc}
\toprule
\multicolumn{1}{c|}{}  & \multicolumn{6}{c|}{\raidar (Ours)} & \multicolumn{3}{c}{\bl{DetectGPT}}  \\
\multicolumn{1}{c|}{LLM Model Used}  & \multicolumn{3}{c|}{ In Distribution} & \multicolumn{3}{c}{ Out of Distribution}\\
\multicolumn{1}{c|}{for Text Generation}  & Code & Yelp & arXiv & Code & Yelp & arXiv & Code & Yelp & arXiv  \\
\midrule
Ada & {96.88} & \textbf{96.15} & {97.10} & 62.06 & {72.72} & {70.00} & \bl{67.39} & \bl{70.59} & \bl{69.74} \\
Text-Davinci-002 & 84.85 & 65.80 & 76.51 & 75.41 & 51.06 & 60.00 & \bl{66.82} & \bl{71.36} & \bl{66.67}\\
        GPT-3.5-turbo & 95.38 & 87.75 & 81.94 & \textbf{91.43} & 71.42 & 48.74 & \bl{67.39} & \bl{69.23} & \bl{66.67} \\
\bl{GPT-4-turbo} & \bl{80.00} & \bl{83.42} & \bl{84.21} & \bl{83.07} & \bl{79.73} & \bl{74.02} & \bl{70.97} & \bl{66.94} & \bl{66.99} \\
\bl{LLaMA 2} & \bl{\textbf{98.46}} & \bl{89.31} & \bl{\textbf{97.87}} & \bl{70.96} & \bl{\textbf{89.30}} & \bl{\textbf{74.41}} & \bl{68.42} & \bl{67.24} & \bl{66.67}\\
\bottomrule
\end{tabular}
\end{center}
\end{table}



\begin{table}[t]
\caption{
Effectiveness of detection using various large language models for rewriting. We present detection F1 scores for the same input data rewritten by Ada, Text-Davinci-002, and GPT-3.5. Among these, GPT-3.5-turbo yields the highest performance in rewriting for detection.
}
\label{tab:diffdetect}
\begin{center}
\begin{tabular}{l|cccccc}
\toprule

\multicolumn{1}{c|}{LLM for}  & \multicolumn{6}{c}{ Datasets} \\
\multicolumn{1}{c|}{Rewriting}  & News & Creative Writing & Student Essay & Code  & \bl{Yelp} & \bl{Arxiv} \\

\midrule
Ada & 55.73 & 62.50 & 57.02 & 77.42 & \bl{73.33} & \bl{71.75} \\
Text-Davinci-002 & 55.47 & 60.59 & 58.96 & 82.19 & \bl{75.15} & \bl{59.25} \\
GPT 3.5 turbo & \textbf{60.29} & \textbf{62.88} & \textbf{64.81} & \textbf{95.38} & \textbf{87.75} & \textbf{81.94} \\
\bl{LLaMA 2} & \bl{56.26} & \bl{61.88} & \bl{60.48} & \bl{85.33} & \bl{74.85} & \bl{72.59} \\
\bottomrule
\end{tabular}
\end{center}
\vspace{-5mm}
\end{table}

\textbf{Detection Robustness against Rephrased Text Generation to Evade Detection.} \citet{krishna2023paraphrasing, sadasivan2023can} show that paraphrasing can often evade detection. In Table~\ref{tab:main1}, we show that our approach can detect GPT text when they are not adversarially rephrased. However, a sophisticated adversary might craft prompts for GPT such that the resulting text, when rewritten, undergoes significant changes, thereby evading our detection. We modify the GPT input using the following rephrases:
\begin{lstlisting}[breakatwhitespace=true]
1. Help me rephrase it in human style
2. Help me rephrase it, so that another GPT rewriting will cause a lot of modifications
\end{lstlisting}
Table~\ref{tab:adaptive_attack} reveals that while our detector, trained on the default single prompt data, can be bypassed by adversarial rephrasing (left columns).  In the right columns, we show results when trained on two of the prompts and tested on the remaining prompts. The detectors are trained on multi-prompt data, which enhances its robustness. Even when tested against unseen adversarial prompts, our detector still identifies machine-generated content designed to elude it, achieving up to 93 points on F1 score. One exception is on the Yelp dataset; the ``no adaptive prompt'' has lower performance on ``multiple training prompts'' than ``single training prompts''. We suspect it is due to the Yelp dataset introducing a larger data difference when prompted differently, and this ``multiple training prompts'' setup will decrease performance due to training and testing on different prompts. In general, results in Table~\ref{tab:adaptive_attack} demonstrate that with proper training, our method can be still robust under rephrased text to evade detection, underscoring the significance of diversifying prompt types when learning our detector.



\textbf{Source of Generated Data.} In our main experiment, we train our detector on text generated from GPT-3.5. We study if our model can still detect machine-generated text when they are generated from a different language model. In Table~\ref{tab:diffsource}, we conduct experiments on text generated from Ada, text-davinci-002, and GPT-3.5 model. For all experiments, we use the same GPT-3.5 to rewrite. 

For in-distribution experiments, we train the detector on data generated from the respective language model. Despite all rewrites being from GPT-3.5, we achieved up to 96 F1 score points. Notably, GPT-3.5 excels at detecting Ada-generated content, indicating our method's versatility in identifying both low (Ada) and high-quality (GPT-3.5) data, even they are generated from a different model. We also evaluate our detection efficiency on the Claude~\citep{claude} generated text on student essay~\citep{verma2023ghostbuster}, where we achieve an F1 score of 57.80.

In the out-of-distribution experiment, we train the detector on data from two language models, assuming it is unaware that the test text will be generated from the third model. Despite a  performance drop on detecting the out-of-distribution test data generated from the third model, our method remains effective in detecting content from this unseen model, underscoring our approach's robustness and adaptability, with up to 91 points on F1 score.




\begin{wrapfigure}{r}{0.4\textwidth}
  \includegraphics[width=1\linewidth]{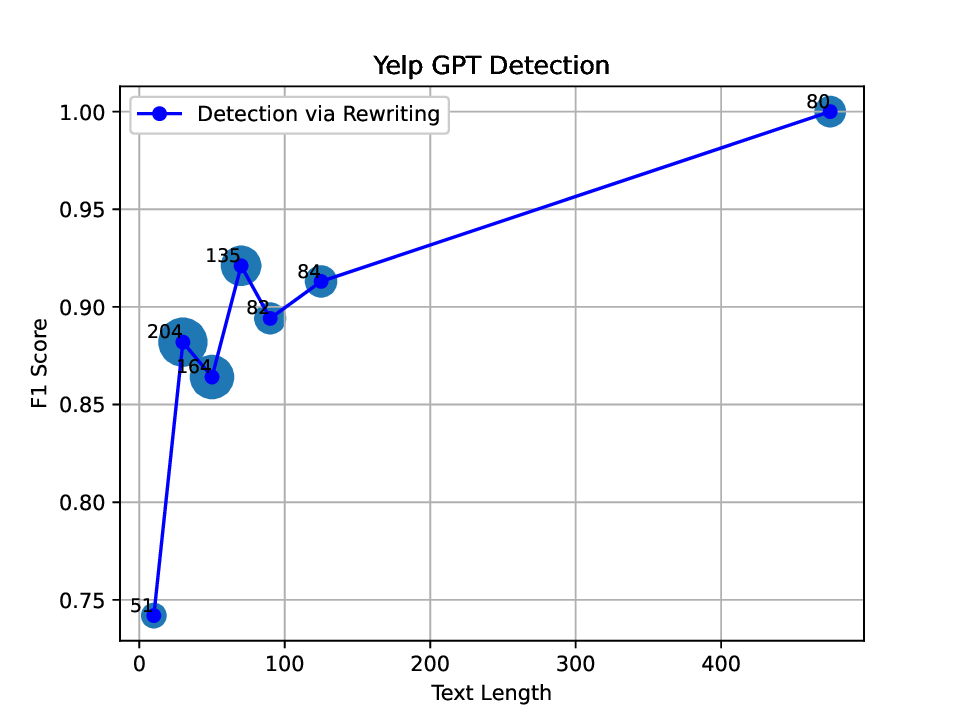}
  \caption{Detection performance as input length increases. On the Yelp dataset, we show that longer input often enables better detection performance. \bl{The number shows the number of data, reflecting by the size of the dot.}}  \label{fig:len}
\end{wrapfigure}

\textbf{Type of Detection Model.} \cite{mireshghallah2023smaller} showed that model size affects performance in perturbation-based detection methods. Given the same input text generated from GPT-3.5, We explore our approach's efficacy with alternative rewriting models with different size. In addition to using the costly GPT-3.5 to rewrite, we incorporate two smaller models, Ada and Text-Davinci-002, and evaluate their detection performance when they are used to rewrite. In Table~\ref{tab:diffdetect}, while all models achieve significant detection performance, our results indicate that a larger rewriting language model enhances detection performance in our method.



\begin{figure*}[t]
\vspace{-5mm}
\centering
\subfloat[Code]{\label{aasxs}\includegraphics[width=0.33\textwidth]{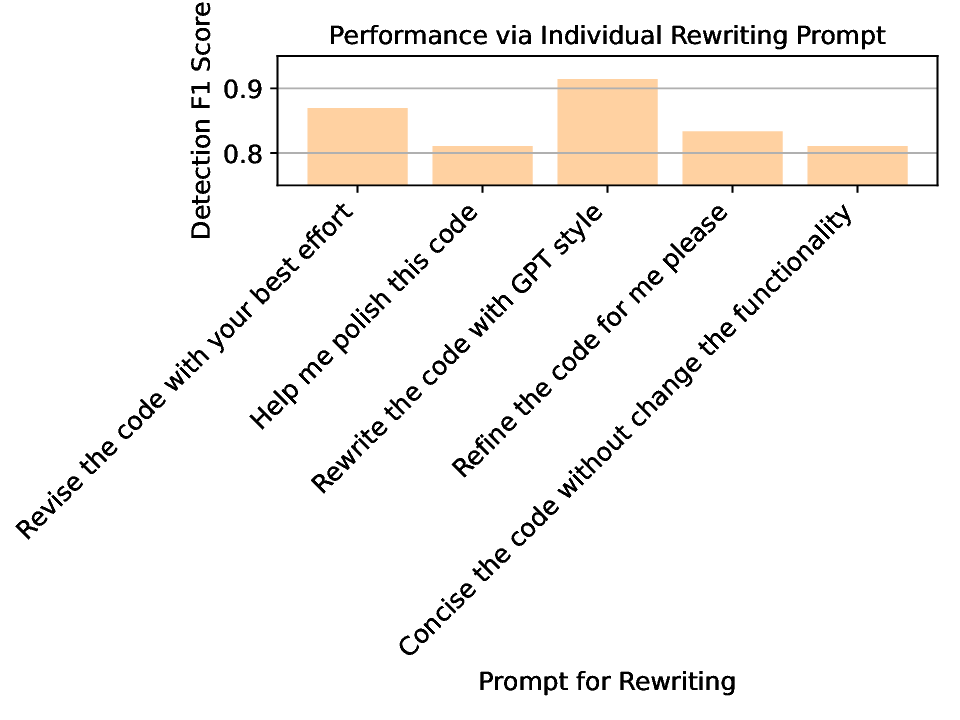}}
\subfloat[Yelp]{\label{aasxs}\includegraphics[width=0.33\textwidth]{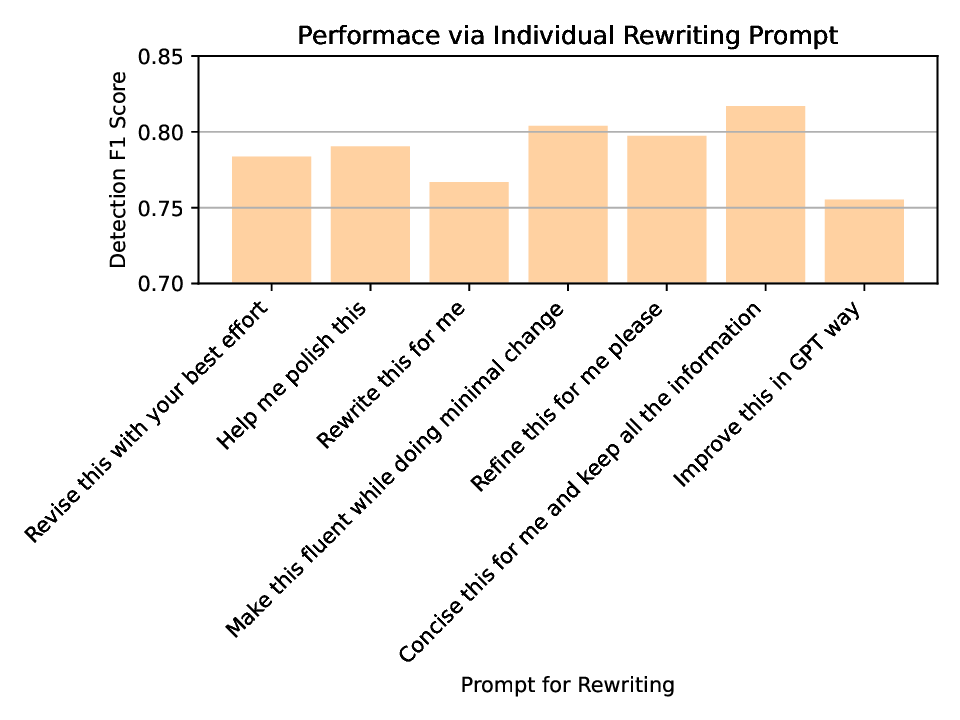}}
\subfloat[ArXiv]{\label{aasxs}\includegraphics[width=0.33\textwidth]{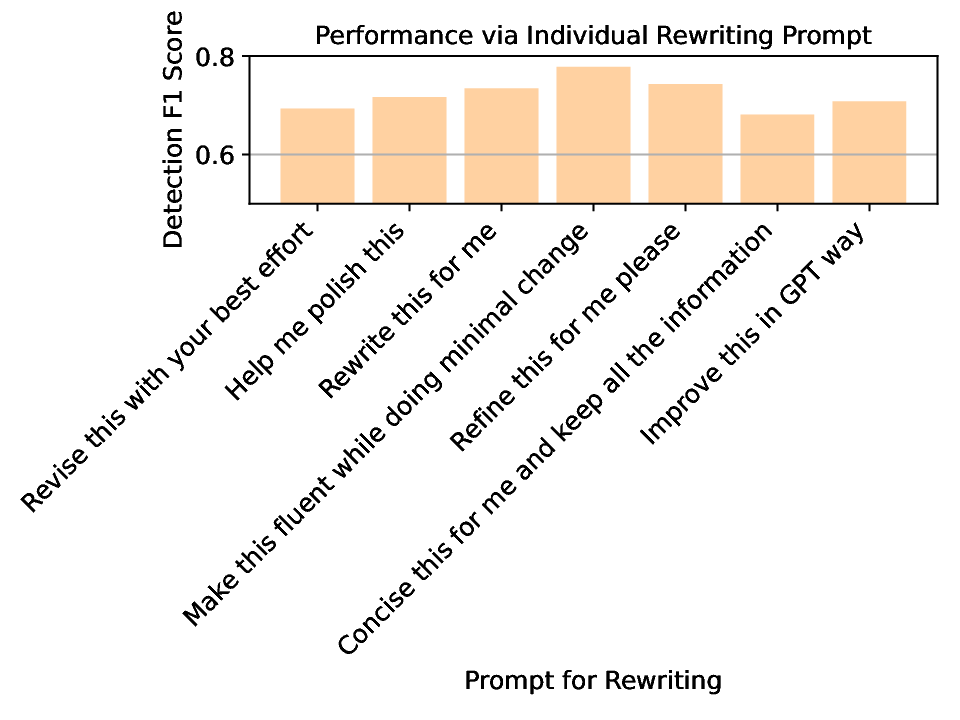}}
\caption{Performance of individual prompt. Different prompts used during rewriting can have a significant impact on the final detection performance. There is no single prompt that performs best across all data sources. With a single rewriting prompt, we can obtain up to 90 points of detection F1 score. } 
\label{fig:prompt}
\vspace{-5mm}
\end{figure*}

\textbf{Impact of Different Prompts.} Figure~\ref{fig:prompt} displays the detection F1 score for various prompts across three datasets. While \citet{mitchell2023detectgpt} employs up to 100 perturbations to query LLM and compute curvature from loss, our approach achieves high detection performance using just a single rewriting prompt.


\textbf{Impact of Content Length.} We assess our detection method's performance across varying input lengths using the Yelp Review dataset in Figure~\ref{fig:len}. Longer inputs, in general, achieve higher detection performance. Notably, while many algorithms fail with shorter inputs~\citep{gptzero, verma2023ghostbuster}, our method can achieve 74 points of detection F1 score even with inputs as brief as ten words, highlighting the effectiveness of our approach.













%% file: files/conclusion.tex
\vspace{-3mm}
\section{Conclusion}
\vspace{-2mm}
We introduce \raidar, an approach to use rewriting editing distance to detect machine-generated text. Our results demonstrate improved detection performance across several benchmarks and state-of-the-art detection methods. Our method is still effective when detecting text generated from novel language models and text generated via prompts that aim to bypass our detection. Our findings show that integrating the inherent structure of large language models can provide useful information to detect text generated from those language models, opening up a new direction for detecting machine-generated text.

%% file: files/appendix.tex
\newpage
\section{Appendix}

\subsection{Data Creation}

\bl{Our dataset selection was driven by the need to address emerging challenges and gaps in current research. We incorporated news, creative writing, and essays from the established Ghostbuster~\cite{verma2023ghostbuster} to maintain continuity with prior work. Recognizing the growing capabilities of Language Learning Models (LLMs) like ChatGPT in generating code, and the accompanying security issues, we included code as a novel and pertinent text data type. Additionally, we analyzed Yelp reviews to explore LLMs' potential for generating fake reviews, a concern overlooked in previous studies, which could significantly influence public opinion about businesses. Lastly, we included arXiv data to address recent concerns over the use of GPT in academic writing, reflecting on its ethical implications.}

\bl{
\textbf{Code Dataset.} Human eval dataset offers code specification and the completed code for each data point. We first use GPT to generate a detailed description of the function of the code by prompting it with “Describe what this code does {code specification}{code}”. The result, termed {pseudo code}, is an interpretation of the code. Subsequently, we prompt GPT with "I want to do this {pseudo code}, help me write code starting with this {code specification}," to generate Python code that adheres to the given input-output format and specifications. This way, we create the AI-generated code data.}

\bl{
\textbf{Yelp Reviews Dataset.} When tasked with crafting a synthetic Yelp review, prompting GPT-3.5 with "Help me write a review based on this {original review}" resulted in verbose and lengthy text. However, we discovered that using the prompt "Write a very short and concise review based on this: {original review}" yielded the most effective and succinct AI-generated reviews.}

\bl{
\textbf{ArXiv Dataset.} In our experiment with Arxiv data, which includes titles and abstracts, we synthesized abstracts by using the title and the first 15 words of the original abstract. We employed the prompt “The title is {title}, start with {first 15 words}, write a short concise abstract based on this:”, which successfully generated realistic abstracts that align with the titles.}

\subsection{Dataset Statistics}

In Table~\ref{tab:sta1} and Table~\ref{tab:sta2}, we show each dataset's size, median, min, and max length on human-written and machine-generated ones, respectively.

\begin{table}[h]
\caption{
\bl{Statistics for each dataset from humans. We show the length in word count. Our work focuses on detecting paragraph-level text, which generally has a shorter and more challenging length.}
}
\label{tab:sta1}
\begin{center}
\begin{tabular}{l|llllll}
\toprule

\multicolumn{1}{c|}{}  & \multicolumn{6}{c}{ Datasets} \\
\multicolumn{1}{c|}{}  & News & Creative Writing & Student Essay & Code  & Yelp & Arxiv \\

\midrule
\bl{Dataset Size} & 730 & 973 & 22172 & 164 & 2000& 350 \\
\bl{Median Length} & 38 & 21 & 96 & 96 & 21 & 102 \\
\bl{Minimum Length} & 2 & 2 & 16 & 2 & 6 & 19\\
\bl{Maximum Length} & 122 & 295 & 1186 & 78 & 1006 & 274\\
\bottomrule
\end{tabular}
\end{center}
\end{table}

\begin{table}[h]
\caption{
\bl{Statistics for each dataset generated by GPT-3.5-Turbo. We show the length in word count. Our work focuses on detecting paragraph-level text, which generally has a shorter and more challenging length.}
}
\label{tab:sta2}
\begin{center}
\begin{tabular}{l|llllll}
\toprule

\multicolumn{1}{c|}{}  & \multicolumn{6}{c}{ Datasets} \\
\multicolumn{1}{c|}{}  & News & Creative Writing & Student Essay & Code  & Yelp & Arxiv \\

\midrule
\bl{Dataset Size} & 479 & 728 & 13629 & 164 & 2000 & 350\\
\bl{Median Length} & 45 & 38 & 82 &  35 & 48 & 72 \\
\bl{Minimum Length} & 3 & 2 & 2 & 5 & 2 & 15\\
\bl{Maximum Length} &  208 & 354 & 291 & 182 & 227 & 129\\
\bottomrule
\end{tabular}
\end{center}
\end{table}




\subsection{Algorithm}\label{sec:alg}

We show the algorithm for invariance, equivariance, and uncertainty based algorithms. We denote the learned classifier as $C$.

\begin{algorithm}[h]
\caption{Detecting LLM Generated Content via Output Invariance}
\label{algorithm: SSLattack}
\begin{algorithmic}[1]
\STATE {\bfseries Input:} Text input $\x$, rephrase prompt $\P_k$, where $k=1, ..., K$.
\STATE {\bfseries Output:} Class prediction $\hat{y}$
\STATE{\bfseries Inference:}

\FOR{$k=1,...,K$}
\STATE{Obtain LLM output $S_k = F(\P_k, \x)$}
\STATE{Calculate bag-of-words edit $R_k$ and the Levenshtein Score $D_k$}
\ENDFOR
\STATE{Make final prediction via $y = C([R_1, R_2, ..., R_K, D_1, D_2, ..., D_K])$}

\end{algorithmic}
\end{algorithm}

\begin{algorithm}[h]
\caption{Detecting LLM Generated Content via Output Equivariance}
\label{algorithm: SSLattack}
\begin{algorithmic}[1]
\STATE {\bfseries Input:} Text input $\x$.
\STATE {\bfseries Output:} Class prediction $\hat{y}$
\STATE{\bfseries Inference:}

\FOR{$k=1,...,K$}
\STATE{Create transformation prompt $\T_k$ and inverse transformation prompt $\T'_k$, create rephrase prompt $P_k$.}
\STATE{Obtain LLM output $\M_k = F(\T_k, \x)$}
\STATE{Obtain LLM output $\M'_k = F(\P_k, \M_k)$}
\STATE{Obtain LLM output $S_k = F(\T'_k, \M'_k)$}
\STATE{Calculate bag-of-words edit $R_k$ and the Levenshtein Score $D_k$}
\ENDFOR
\STATE{Make final prediction via $y = C([R_1, R_2, ..., R_K, D_1, D_2, ..., D_K])$}

\end{algorithmic}
\end{algorithm}

\begin{algorithm}[h]
\caption{Detecting LLM generated Content via Output Uncertainty}
\label{algorithm: SSLattack}
\begin{algorithmic}[1]
\STATE {\bfseries Input:} Text input $\x$.
\STATE {\bfseries Output:} Class prediction $\hat{y}$
\STATE{\bfseries Inference:}
\STATE{Given rephrase prompt $\P$}
\FOR{$k=1,...,K$}
\STATE{Obtain LLM output $S_k = F(\P, \x)$}
\ENDFOR
\FOR{$k=1,...,K$}
\FOR{$j=k,...,K$}
\STATE{Calculate bag-of-words edit $R_{k,j}$ and the Levenshtein Score $D_{k,j}$}
\ENDFOR
\ENDFOR
\STATE{Make final prediction via $y = C([R_{1,2}, R_{1,3}, ..., R_{K-1, K}, D_{1,2}, D_{1,3}, ..., R_{K-1, K}])$}

\end{algorithmic}
\end{algorithm}

\begin{figure*}[t]
\vspace{-9mm}
\centering
\subfloat[News]{\label{aasxs}\includegraphics[width=0.49\textwidth]{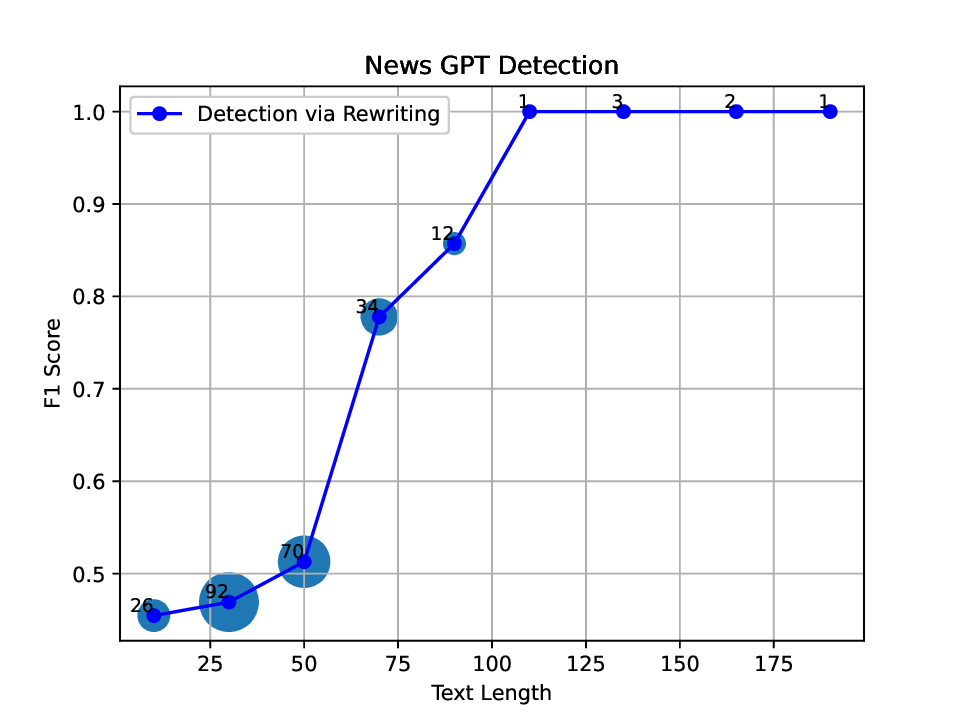}}
\subfloat[Creative Writing]{\label{aasxs}\includegraphics[width=0.49\textwidth]{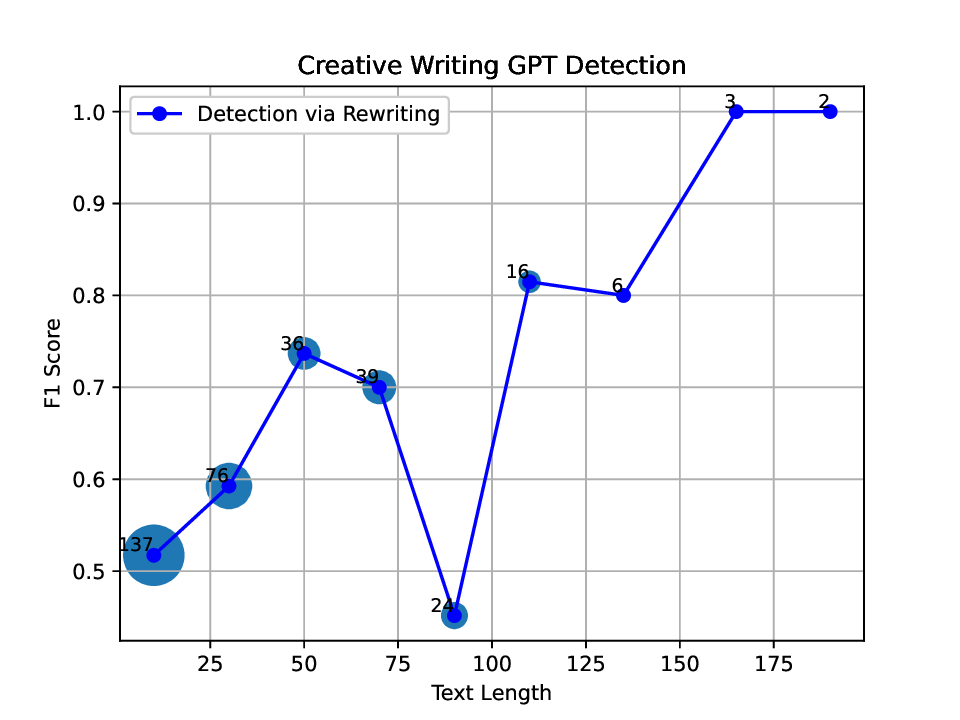}}
\caption{\bl{Detection performance under different length. For News and Creative Writing datasets, longer length helps detection.}} 
\label{fig:len1}
\end{figure*}

\begin{figure*}[t]
\centering
\subfloat[Student Essay]{\label{aasxs}\includegraphics[width=0.49\textwidth]{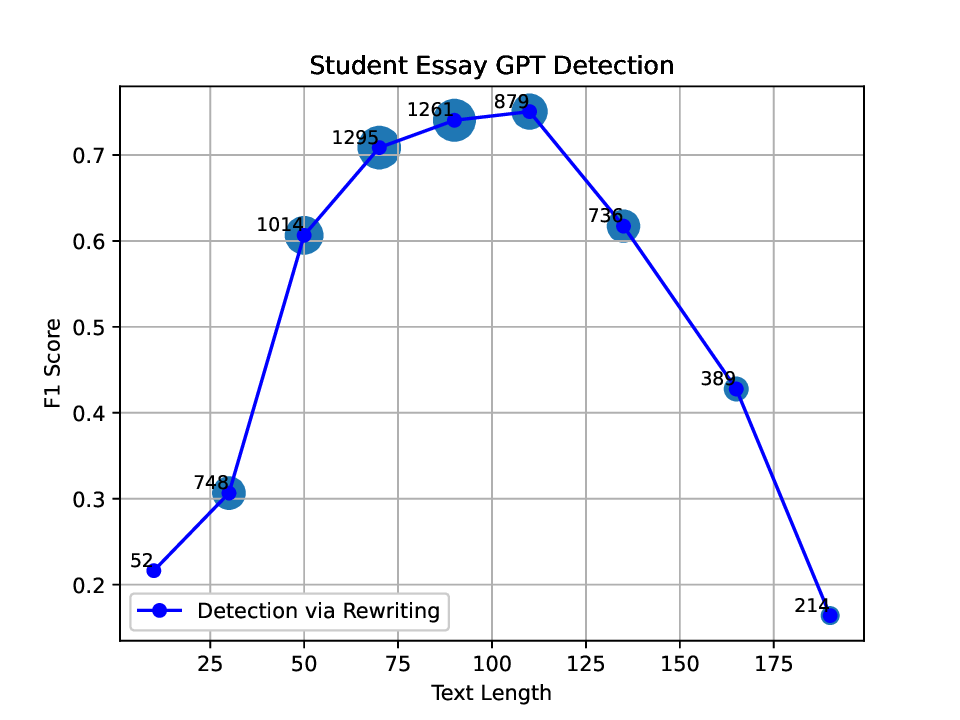}}
\subfloat[Yelp Reviews]{\label{aasxs}\includegraphics[width=0.49\textwidth]{Figs/len_ablation.eps}}
\caption{\bl{Detection performance under different length. For both datasets, longer length helps detection. Yet, on student essay, input longer than 125 words will lead to performance degradation.}} 
\label{fig:len2}
\end{figure*}

\begin{figure*}[t]
\centering
\subfloat[Code]{\label{aasxs}\includegraphics[width=0.49\textwidth]{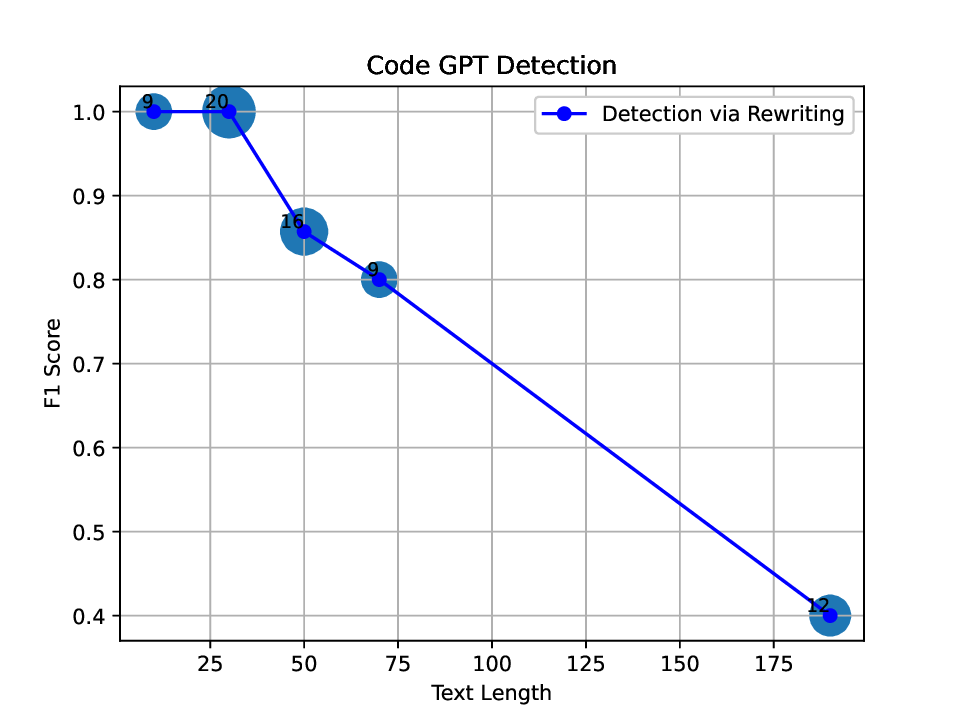}}
\subfloat[Arxiv Abstract]{\label{aasxs}\includegraphics[width=0.49\textwidth]{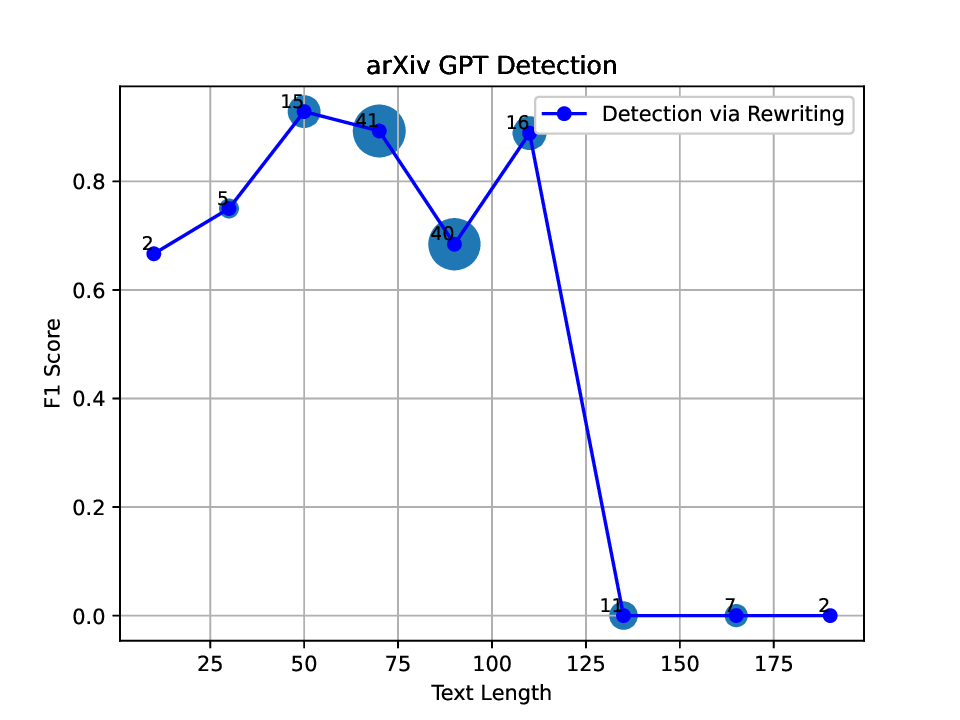}}
\caption{\bl{Detection performance under different length. For both datasets, the performance is high in the beginning, demonstrating the advantage of our approach in tackling sequence that is shorter. However, for longer input, the detection performance drops.}} 
\label{fig:len3}
\end{figure*}

\subsection{Analysis}
\bl{
\textbf{Quality of the Machine-Generated Content.} LLM tends to treat the text generated by the machine as high quality and conducts few edits. We conduct a human study on whether the text generated by machines is indeed of higher quality than that written by humans. This study focused on the Yelp and Arxiv datasets. Participants were presented with two pieces of text designed for the same purpose, one authored by a human and the other by a machine, and were asked to judge which was of higher quality. The study involved three users, and for each dataset, we randomly selected 20 examples for evaluation. The results, detailed in Table~\ref{tab:humanquality}, generally indicate that human-written texts are of similar or higher quality compared to those generated by machines.}

\begin{table}[h]
\caption{\bl{Human study on the quality of machine generated text. Our work showed that machine generated text will be perferred by LLMs and produce few edits when asked to rewrite. We also evaluate the ratio of machine generated text that is perferred by human. The machine is good at creating realistic Yelp reviews, but not good at academia paper writing.}}
\label{tab:humanquality}
\begin{center}
\begin{tabular}{l|cc}
\toprule
\multicolumn{1}{c|}{Methods}  & Yelp  &  Arxiv    \\
\multicolumn{1}{c|}{}  &  Reviews &   Abstract   \\

\midrule

\% that Machine Generated Text are Preferred Human Written Text & 53.3\% & 26.7\% \\
\bottomrule
\end{tabular}
\end{center}
\end{table}

\bl{\textbf{Robustness of Our Method to LLM Fine-tuning.}
We run the experiment on GPT-3.5-Turbo and GPT-4-Turbo. GPT-4-Turbo can be roughly treated as a realistic, advanced, continual fine-tuned LLM on new real-world data from GPT-3.5-Turbo. We show the results in Table~\ref{tab:finetune}. Our method is robust to LLM finetuned. Despite a drop in detection performance, it still outperforms the established state-of-the-art zero-shot detector.}

\begin{table}[h]
\caption{
\bl{Robustness of Our Algorithm to LLM Finetuning. The detection model was only learned on GPT-3.5-Turbo generated data and use GPT-3.5-Turbo for rewriting. We show the results on GPT-3.5-Turbo in the first row. We then directly apply the detector to data generated from GPT-4-Turbo, but use the old, GPT-3.5-Turbo model for rewriting and detection. The detector was never trained on GPT-4-Turbo. Despite a drop in detection effectiveness, our algorithm still outperform the published state-of-the-art zero-shot detector.}
}
\label{tab:finetune}
\begin{center}
\begin{tabular}{l|l|ccc}
\toprule
 & & \multicolumn{3}{c}{ Datasets} \\
\multicolumn{1}{c|}{Test Data source} & \multicolumn{1}{c|}{Data Detector} & Code  & Yelp & Arxiv \\
\midrule
GPT-3.5-Turbo & Ours trained with GPT-3.5-Turbo & 95.38 & 87.75 & 81.94  \\
\cellcolor{Gray}GPT-4-Turbo & \cellcolor{Gray}Ours trained with GPT-3.5-Turbo &  \cellcolor{Gray}83.07 & \cellcolor{Gray}79.73 & \cellcolor{Gray}74.02 \\
\midrule
GPT-4-Turbo & Baseline DetectGPT &  70.97 & 66.94 & 66.99 \\
\bottomrule
\end{tabular}
\end{center}
\end{table}

\bl{
\textbf{Robustness of Our Method to Non-native Speaker.} Prior work showed that LLM detectors are biased against non-native English writers, because non-native English writing is limited in linguistic expressions and is often detected as AI-generated~\cite{liang2023gpt}. We investigate if our approach can detect non-native English writers better or if it is biased against them, as shown by prior detection methods. }

\bl{
Following the setup from Liang et al~\cite{liang2023gpt}, we use the Hewlett Foundation's Automated Student Assessment Prize (ASAP) dataset and adopt the first 200 datasets in our study, which is a dataset from non-native speakers on TOEFL essays on 8-th grade level in the US. We create the machine-generated answer for the TOEFL essay via the following prompt:}
\begin{lstlisting}[breakatwhitespace=true]
Write an essay based on this:
\end{lstlisting}

\bl{
We show the detection result in Table~\ref{tab:non-native}. Our method does not discriminate the non-English speaker, and reaches a similar level of detection performance on high-quality writing (abstract from accepted ICLR papers). Since both ASAP and Arxiv are written by humans, they will be treated as low-quality text that does not match the inherent inertia in LLM models, and thus will both be modified more than the machine-generated text. Our detection algorithm will classify those texts with more modifications than humans. Thus, both non-native and efficient writers will be correctly classified by our approach. Since our algorithm only relies on the word edit distance, it does not rely on the superficial semantics of the text for detection. Thus, our approach generalizes well from academic ICLR abstract to non-native English writing on the 8th grade level, with only less than 1 point of performance drop.}

\begin{table}[t]
\caption{\bl{Robustness on non-native English authors. We show results that train on our Arxiv dataset, and test on the ASAP dataset in the gray row. While the detection score drop a bit from training on ASAP and test on ASAP, we still achieve a F-1 detection score of 81.16, which is only less than 1 point than Arxiv paper. This demonstrate the robustness of our detection algorithm, even trained on the Arxiv papers that are accepted to ICLR, which are high quality written text, our algorithm still generalize well to non-native English writters from grade 8 level.}}
\label{tab:non-native}
\begin{center}
\begin{tabular}{l|cc}
\toprule
Training Source  & Testing Source & F-1 Detection Score \\
\midrule
ASAP Dataset & ASAP Dataset& 98.76  \\
\cellcolor{Gray}Arxiv Dataset& \cellcolor{Gray}ASAP Dataset& \cellcolor{Gray}81.16  \\
Arxiv Dataset& Arxiv Dataset& 81.95  \\
\bottomrule
\end{tabular}
\end{center}
\end{table}




\bl{
\textbf{Detection Performance by combining rewrites from multiple LLMs.} In Table~\ref{tab:combineLLMs}, we show detection performance when combining GPT-3.5 rewrites with other LLMs, including Ada, Davinci, and both. We find combining rewriting from multiple LLMs can improve performance over Arxiv detection, but not on Yelp.}

\begin{table}[h]
\caption{\bl{F1 score for detecting machine-generated paragraphs by combining rewrites from multiple LLMs. We experiment on two datasets.}}
\label{tab:combineLLMs}
\begin{center}
\begin{tabular}{l|cc}
\toprule
\multicolumn{1}{c|}{Methods}  & Yelp Reviews &  Arxiv Abstract   \\

\midrule

 GPT-3.5 Only & \cellcolor{Gray}\textbf{87.75} & \cellcolor{Gray}81.94  \\
 GPT-3.5 + Ada & 85.71 & \textbf{92.85} \\
 GPT-3.5 + Davinci-003 & 85.53 & 88.40 \\
 GPT-3.5 + Davinci-003 + Ada & 81.76 &  90.00 \\
\bottomrule
\end{tabular}
\end{center}
\end{table}

\bl{
\textbf{Detection performance by adding edit distance between the rewritten texts from different LLMs as additional features.} In Table~\ref{tab:diffbetweenLLM}, we show the detection performance. We can achieve better detection performance leveraging this new feature.}

\begin{table}[h]
\caption{\bl{F1 score for detecting machine-generated paragraphs by edit distance between the rewritten texts from different LLMs as additional features.}}
\label{tab:diffbetweenLLM}
\begin{center}
\begin{tabular}{l|cc}
\toprule
\multicolumn{1}{c|}{Methods}  & Yelp Reviews &  Arxiv Abstract   \\

\midrule

 GPT-3.5 Only & \cellcolor{Gray}\textbf{87.75} & \cellcolor{Gray}81.94  \\
 GPT-3.5 / Ada & 67.85 & 89.21 \\
 GPT-3.5 / Davinci & 78.41 & 81.94 \\
 Ada / Davinci & 66.25 & \textbf{90.51}\\
\bottomrule
\end{tabular}
\end{center}
\end{table}

\bl{
\textbf{Detection performance by combining features of invariance, equivariance, and uncertainty.} We conduct experiments in Table~\ref{tab:combine3}, on the two dataset we studied, we cannot further improve performance.}

\bl{
\textbf{Detection performance under different input length.} We show the trend in Figure~\ref{fig:len1}, Figure~\ref{fig:len2}, and Figure~\ref{fig:len3}.}

\begin{table}[h]
\caption{\bl{Detection performance combining invariance, equivariance, and uncertainty.}}
\label{tab:combine3}
\begin{center}
\begin{tabular}{l|cc}
\toprule
\multicolumn{1}{c|}{Methods}  & News &  Creative Writing   \\

\midrule

 Single & \textbf{60.29} & \textbf{62.88}\\
 Combined & 53.72 & 58.18  \\
\bottomrule
\end{tabular}
\end{center}
\end{table}

\bl{
\textbf{Statistical significance of the number of changes (deletions, insertions) done by the selective generative models between humans and machine-generated texts.} We calculate the t-statistic and calculate the p-value. In Table~\ref{tab:tvalue}, we show the p-value for the two distributions shown in Figure~\ref{fig:curve}. Since the p-value is much smaller than 0.05, it demonstrates that the number of changes between human and machine-generated text is significant.}

\begin{table}[h]
\caption{\bl{Statistical significance of the number of changes (deletions, insertions) done by the selective generative models between humans and machine-generated texts, corresponding to Figure 2 in the main paper. We present the p value by running a one-sided two-sample t tests. The small p-value demonstrates the statistical significance.}}
\label{tab:tvalue}
\begin{center}
\begin{tabular}{l|ccc}
\toprule
\multicolumn{1}{c|}{Methods}  & Invariance &  Equivariance & Uncertainty   \\
\midrule
 p-value & 2.19e-13 & 9.21e-7 & 5.47e-16\\
\bottomrule
\end{tabular}
\end{center}
\end{table}

\subsection{Implementation Details}

\bl{
\textbf{The training and testing domain for Table 2.}
For all experiments in Table 2, we use logistic regression, and use the same source and target for invariance, equivariance, and uncertainty.
For News, we train on Creative Writing and test on News.
For Creative Writing, we train on News and test on Creative Writing.
FOr Student Essay, we train on News, and test on student Essay.}

\bl{
\textbf{Classifier choice for Table 1 and Table 2.} We use logistic regression for all our experiments except for on the student essay dataset, where we find XGBoost achieves better performance.}

\begin{table}[t]
\vspace{-5mm}
\caption{\bl{Classifier for detecting machine-generated paragraphs. We use the best classifier from logistic regression (LR) and XG Boost for classification.}}
\vspace{-5mm}
\label{tab:main1}
\begin{center}
\begin{tabular}{l|cccccc}
\toprule
\multicolumn{1}{c|}{}  & \multicolumn{6}{c}{ Datasets} \\
\multicolumn{1}{c|}{} & & Creative & Student & & Yelp & Arxiv \\
\multicolumn{1}{c|}{Methods}  & News &  Writing &  Essay & Code &  Reviews &  Abstract   \\
\midrule
Ours Invariance & LR & LR & XGBoost & LR & LR & LR \\
Ours Equivariance & LR & LR & XGBoost & LR & LR & LR\\
Ours Uncertainty & LR & LR & XGBoost & LR & LR & LR \\
\bottomrule
\end{tabular}
\end{center}
\end{table}

%% file: iclr2024_conference.bbl
\begin{thebibliography}{45}
\providecommand{\natexlab}[1]{#1}
\providecommand{\url}[1]{\texttt{#1}}
\expandafter\ifx\csname urlstyle\endcsname\relax
  \providecommand{\doi}[1]{doi: #1}\else
  \providecommand{\doi}{doi: \begingroup \urlstyle{rm}\Url}\fi

\bibitem[Cha(2023)]{ChatGPT}
Chatgpt: Optimizing language models for dialogue, 2023.
\newblock URL \url{https://chat.openai.com}.

\bibitem[Anthropic(2023)]{claude}
Anthropic, 2023.
\newblock URL \url{https://www.anthropic.com/product}.

\bibitem[Asfour \& Murillo(2023)Asfour and Murillo]{asfour2023harnessing}
Mohammad Asfour and Juan~Carlos Murillo.
\newblock Harnessing large language models to simulate realistic human
  responses to social engineering attacks: A case study.
\newblock \emph{International Journal of Cybersecurity Intelligence \&
  Cybercrime}, 6\penalty0 (2):\penalty0 21--49, 2023.

\bibitem[Bakhtin et~al.(2019)Bakhtin, Gross, Ott, Deng, Ranzato, and
  Szlam]{bakhtin2019real}
Anton Bakhtin, Sam Gross, Myle Ott, Yuntian Deng, Marc'Aurelio Ranzato, and
  Arthur Szlam.
\newblock Real or fake? learning to discriminate machine from human generated
  text.
\newblock \emph{arXiv preprint arXiv:1906.03351}, 2019.

\bibitem[Bergman et~al.(2022)Bergman, Abercrombie, Spruit, Hovy, Dinan,
  Boureau, Rieser, et~al.]{bergman2022guiding}
A~Stevie Bergman, Gavin Abercrombie, Shannon Spruit, Dirk Hovy, Emily Dinan,
  Y-Lan Boureau, Verena Rieser, et~al.
\newblock Guiding the release of safer e2e conversational ai through value
  sensitive design.
\newblock In \emph{Proceedings of the 23rd Annual Meeting of the Special
  Interest Group on Discourse and Dialogue}. Association for Computational
  Linguistics, 2022.

\bibitem[Berkson(1944)]{berkson1944application}
Joseph Berkson.
\newblock Application of the logistic function to bio-assay.
\newblock \emph{Journal of the American Statistical Association}, 39\penalty0
  (227):\penalty0 357--365, 1944.

\bibitem[Brown et~al.(2020)Brown, Mann, Ryder, Subbiah, Kaplan, Dhariwal,
  Neelakantan, Shyam, Sastry, Askell, et~al.]{GPT3}
Tom Brown, Benjamin Mann, Nick Ryder, Melanie Subbiah, Jared~D Kaplan, Prafulla
  Dhariwal, Arvind Neelakantan, Pranav Shyam, Girish Sastry, Amanda Askell,
  et~al.
\newblock Language models are few-shot learners.
\newblock \emph{Advances in neural information processing systems},
  33:\penalty0 1877--1901, 2020.

\bibitem[Chakraborty et~al.(2023)Chakraborty, Bedi, Zhu, An, Manocha, and
  Huang]{chakraborty2023possibilities}
Souradip Chakraborty, Amrit~Singh Bedi, Sicheng Zhu, Bang An, Dinesh Manocha,
  and Furong Huang.
\newblock On the possibilities of ai-generated text detection.
\newblock \emph{arXiv preprint arXiv:2304.04736}, 2023.

\bibitem[Chen et~al.(2021)Chen, Tworek, Jun, Yuan, de~Oliveira~Pinto, Kaplan,
  Edwards, Burda, Joseph, Brockman, Ray, Puri, Krueger, Petrov, Khlaaf, Sastry,
  Mishkin, Chan, Gray, Ryder, Pavlov, Power, Kaiser, Bavarian, Winter, Tillet,
  Such, Cummings, Plappert, Chantzis, Barnes, Herbert-Voss, Guss, Nichol,
  Paino, Tezak, Tang, Babuschkin, Balaji, Jain, Saunders, Hesse, Carr, Leike,
  Achiam, Misra, Morikawa, Radford, Knight, Brundage, Murati, Mayer, Welinder,
  McGrew, Amodei, McCandlish, Sutskever, and Zaremba]{chen2021codex}
Mark Chen, Jerry Tworek, Heewoo Jun, Qiming Yuan, Henrique~Ponde
  de~Oliveira~Pinto, Jared Kaplan, Harri Edwards, Yuri Burda, Nicholas Joseph,
  Greg Brockman, Alex Ray, Raul Puri, Gretchen Krueger, Michael Petrov, Heidy
  Khlaaf, Girish Sastry, Pamela Mishkin, Brooke Chan, Scott Gray, Nick Ryder,
  Mikhail Pavlov, Alethea Power, Lukasz Kaiser, Mohammad Bavarian, Clemens
  Winter, Philippe Tillet, Felipe~Petroski Such, Dave Cummings, Matthias
  Plappert, Fotios Chantzis, Elizabeth Barnes, Ariel Herbert-Voss,
  William~Hebgen Guss, Alex Nichol, Alex Paino, Nikolas Tezak, Jie Tang, Igor
  Babuschkin, Suchir Balaji, Shantanu Jain, William Saunders, Christopher
  Hesse, Andrew~N. Carr, Jan Leike, Josh Achiam, Vedant Misra, Evan Morikawa,
  Alec Radford, Matthew Knight, Miles Brundage, Mira Murati, Katie Mayer, Peter
  Welinder, Bob McGrew, Dario Amodei, Sam McCandlish, Ilya Sutskever, and
  Wojciech Zaremba.
\newblock Evaluating large language models trained on code.
\newblock 2021.

\bibitem[Chen \& Guestrin(2016)Chen and Guestrin]{chen2016xgboost}
Tianqi Chen and Carlos Guestrin.
\newblock Xgboost: A scalable tree boosting system.
\newblock In \emph{Proceedings of the 22nd ACM SIGKDD International Conference
  on Knowledge Discovery and Data Mining}, pp.\  785--794. ACM, 2016.

\bibitem[Chowdhery et~al.(2022)Chowdhery, Narang, Devlin, Bosma, Mishra,
  Roberts, Barham, Chung, Sutton, Gehrmann, et~al.]{chowdhery2022palm}
Aakanksha Chowdhery, Sharan Narang, Jacob Devlin, Maarten Bosma, Gaurav Mishra,
  Adam Roberts, Paul Barham, Hyung~Won Chung, Charles Sutton, Sebastian
  Gehrmann, et~al.
\newblock Palm: Scaling language modeling with pathways.
\newblock \emph{arXiv preprint arXiv:2204.02311}, 2022.

\bibitem[Cotton et~al.(2023)Cotton, Cotton, and Shipway]{cotton2023chatting}
Debby~RE Cotton, Peter~A Cotton, and J~Reuben Shipway.
\newblock Chatting and cheating: Ensuring academic integrity in the era of
  chatgpt.
\newblock \emph{Innovations in Education and Teaching International}, pp.\
  1--12, 2023.

\bibitem[Dou et~al.(2021)Dou, Forbes, Koncel-Kedziorski, Smith, and Choi]{dou}
Yao Dou, Maxwell Forbes, Rik Koncel-Kedziorski, Noah~A Smith, and Yejin Choi.
\newblock Is gpt-3 text indistinguishable from human text? scarecrow: A
  framework for scrutinizing machine text.
\newblock \emph{arXiv preprint arXiv:2107.01294}, 2021.

\bibitem[Fagni et~al.(2021)Fagni, Falchi, Gambini, Martella, and
  Tesconi]{fagni2021tweepfake}
Tiziano Fagni, Fabrizio Falchi, Margherita Gambini, Antonio Martella, and
  Maurizio Tesconi.
\newblock Tweepfake: About detecting deepfake tweets.
\newblock \emph{Plos one}, 16\penalty0 (5):\penalty0 e0251415, 2021.

\bibitem[Gehrmann et~al.(2019)Gehrmann, Strobelt, and Rush]{gehrmann2019gltr}
Sebastian Gehrmann, Hendrik Strobelt, and Alexander~M Rush.
\newblock Gltr: Statistical detection and visualization of generated text.
\newblock \emph{arXiv preprint arXiv:1906.04043}, 2019.

\bibitem[Ippolito et~al.(2019)Ippolito, Duckworth, Callison-Burch, and
  Eck]{ippolito2019automatic}
Daphne Ippolito, Daniel Duckworth, Chris Callison-Burch, and Douglas Eck.
\newblock Automatic detection of generated text is easiest when humans are
  fooled.
\newblock \emph{arXiv preprint arXiv:1911.00650}, 2019.

\bibitem[Jawahar et~al.(2020)Jawahar, Abdul-Mageed, and
  Lakshmanan]{jawahar2020automatic}
Ganesh Jawahar, Muhammad Abdul-Mageed, and Laks~VS Lakshmanan.
\newblock Automatic detection of machine generated text: A critical survey.
\newblock \emph{arXiv preprint arXiv:2011.01314}, 2020.

\bibitem[Jin et~al.(2019)Jin, Jin, Zhou, and Szolovits]{jin2019bert}
Di~Jin, Zhijing Jin, Joey~Tianyi Zhou, and Peter Szolovits.
\newblock Is bert really robust? natural language attack on text classification
  and entailment.
\newblock \emph{arXiv preprint arXiv:1907.11932}, 2019.

\bibitem[Kang et~al.(2023)Kang, Li, Stoica, Guestrin, Zaharia, and
  Hashimoto]{kang2023exploiting}
Daniel Kang, Xuechen Li, Ion Stoica, Carlos Guestrin, Matei Zaharia, and
  Tatsunori Hashimoto.
\newblock Exploiting programmatic behavior of llms: Dual-use through standard
  security attacks.
\newblock \emph{arXiv preprint arXiv:2302.05733}, 2023.

\bibitem[Kirchenbauer et~al.(2023)Kirchenbauer, Geiping, Wen, Katz, Miers, and
  Goldstein]{kirchenbauer2023watermark}
John Kirchenbauer, Jonas Geiping, Yuxin Wen, Jonathan Katz, Ian Miers, and Tom
  Goldstein.
\newblock A watermark for large language models.
\newblock \emph{arXiv preprint arXiv:2301.10226}, 2023.

\bibitem[Kojima et~al.(2022)Kojima, Gu, Reid, Matsuo, and
  Iwasawa]{kojima2022large}
Takeshi Kojima, Shixiang~Shane Gu, Machel Reid, Yutaka Matsuo, and Yusuke
  Iwasawa.
\newblock Large language models are zero-shot reasoners.
\newblock \emph{Advances in neural information processing systems},
  35:\penalty0 22199--22213, 2022.

\bibitem[Krishna et~al.(2023)Krishna, Song, Karpinska, Wieting, and
  Iyyer]{krishna2023paraphrasing}
Kalpesh Krishna, Yixiao Song, Marzena Karpinska, John Wieting, and Mohit Iyyer.
\newblock Paraphrasing evades detectors of ai-generated text, but retrieval is
  an effective defense.
\newblock \emph{arXiv preprint arXiv:2303.13408}, 2023.

\bibitem[Levenshtein(1966)]{levenshtein1966binary}
Vladimir~I Levenshtein.
\newblock Binary codes capable of correcting deletions, insertions, and
  reversals.
\newblock \emph{Soviet Physics Doklady}, 10\penalty0 (8):\penalty0 707--710,
  1966.

\bibitem[Li et~al.(2022)Li, Tang, Zhao, Nie, and Wen]{li2022pretrained}
Junyi Li, Tianyi Tang, Wayne~Xin Zhao, Jian-Yun Nie, and Ji-Rong Wen.
\newblock Pretrained language models for text generation: A survey.
\newblock \emph{arXiv preprint arXiv:2201.05273}, 2022.

\bibitem[Li \& Liang(2021)Li and Liang]{li2021prefix}
Xiang~Lisa Li and Percy Liang.
\newblock Prefix-tuning: Optimizing continuous prompts for generation.
\newblock \emph{arXiv preprint arXiv:2101.00190}, 2021.

\bibitem[Liang et~al.(2023)Liang, Yuksekgonul, Mao, Wu, and Zou]{liang2023gpt}
Weixin Liang, Mert Yuksekgonul, Yining Mao, Eric Wu, and James Zou.
\newblock Gpt detectors are biased against non-native english writers.
\newblock \emph{arXiv preprint arXiv:2304.02819}, 2023.

\bibitem[Mireshghallah et~al.(2023)Mireshghallah, Mattern, Gao, Shokri, and
  Berg-Kirkpatrick]{mireshghallah2023smaller}
Fatemehsadat Mireshghallah, Justus Mattern, Sicun Gao, Reza Shokri, and Taylor
  Berg-Kirkpatrick.
\newblock Smaller language models are better black-box machine-generated text
  detectors.
\newblock \emph{arXiv preprint arXiv:2305.09859}, 2023.

\bibitem[Mirsky et~al.(2022)Mirsky, Demontis, Kotak, Shankar, Gelei, Yang,
  Zhang, Pintor, Lee, Elovici, et~al.]{mirsky2022threat}
Yisroel Mirsky, Ambra Demontis, Jaidip Kotak, Ram Shankar, Deng Gelei, Liu
  Yang, Xiangyu Zhang, Maura Pintor, Wenke Lee, Yuval Elovici, et~al.
\newblock The threat of offensive ai to organizations.
\newblock \emph{Computers \& Security}, pp.\  103006, 2022.

\bibitem[Mitchell et~al.(2023)Mitchell, Lee, Khazatsky, Manning, and
  Finn]{mitchell2023detectgpt}
Eric Mitchell, Yoonho Lee, Alexander Khazatsky, Christopher~D Manning, and
  Chelsea Finn.
\newblock Detectgpt: Zero-shot machine-generated text detection using
  probability curvature.
\newblock \emph{arXiv preprint arXiv:2301.11305}, 2023.

\bibitem[Pan et~al.(2023)Pan, Pan, Chen, Nakov, Kan, and Wang]{pan2023risk}
Yikang Pan, Liangming Pan, Wenhu Chen, Preslav Nakov, Min-Yen Kan, and
  William~Yang Wang.
\newblock On the risk of misinformation pollution with large language models.
\newblock \emph{arXiv preprint arXiv:2305.13661}, 2023.

\bibitem[Pearce et~al.(2022)Pearce, Ahmad, Tan, Dolan-Gavitt, and
  Karri]{pearce2022asleep}
Hammond Pearce, Baleegh Ahmad, Benjamin Tan, Brendan Dolan-Gavitt, and Ramesh
  Karri.
\newblock Asleep at the keyboard? assessing the security of github copilot’s
  code contributions.
\newblock In \emph{2022 IEEE Symposium on Security and Privacy (SP)}, pp.\
  754--768. IEEE, 2022.

\bibitem[Radford et~al.(2019)Radford, Wu, Child, Luan, Amodei, Sutskever,
  et~al.]{radford2019language}
Alec Radford, Jeffrey Wu, Rewon Child, David Luan, Dario Amodei, Ilya
  Sutskever, et~al.
\newblock Language models are unsupervised multitask learners.
\newblock \emph{OpenAI blog}, 1\penalty0 (8):\penalty0 9, 2019.

\bibitem[Radford et~al.(2023)Radford, Kim, Xu, Brockman, McLeavey, and
  Sutskever]{radford2023robust}
Alec Radford, Jong~Wook Kim, Tao Xu, Greg Brockman, Christine McLeavey, and
  Ilya Sutskever.
\newblock Robust speech recognition via large-scale weak supervision.
\newblock In \emph{International Conference on Machine Learning}, pp.\
  28492--28518. PMLR, 2023.

\bibitem[Sadasivan et~al.(2023)Sadasivan, Kumar, Balasubramanian, Wang, and
  Feizi]{sadasivan2023can}
Vinu~Sankar Sadasivan, Aounon Kumar, Sriram Balasubramanian, Wenxiao Wang, and
  Soheil Feizi.
\newblock Can ai-generated text be reliably detected?
\newblock \emph{arXiv preprint arXiv:2303.11156}, 2023.

\bibitem[Shumailov et~al.(2023)Shumailov, Shumaylov, Zhao, Gal, Papernot, and
  Anderson]{shumailov2023curse}
Ilia Shumailov, Zakhar Shumaylov, Yiren Zhao, Yarin Gal, Nicolas Papernot, and
  Ross Anderson.
\newblock The curse of recursion: Training on generated data makes models
  forget.
\newblock \emph{arXiv preprint arxiv:2305.17493}, 2023.

\bibitem[Siddiq et~al.(2022)Siddiq, Majumder, Mim, Jajodia, and
  Santos]{siddiq2022empirical}
Mohammed~Latif Siddiq, Shafayat~H Majumder, Maisha~R Mim, Sourov Jajodia, and
  Joanna~CS Santos.
\newblock An empirical study of code smells in transformer-based code
  generation techniques.
\newblock In \emph{2022 IEEE 22nd International Working Conference on Source
  Code Analysis and Manipulation (SCAM)}, pp.\  71--82. IEEE, 2022.

\bibitem[Solaiman et~al.(2019)Solaiman, Brundage, Clark, Askell, Herbert-Voss,
  Wu, Radford, Krueger, Kim, Kreps, et~al.]{solaiman2019release}
Irene Solaiman, Miles Brundage, Jack Clark, Amanda Askell, Ariel Herbert-Voss,
  Jeff Wu, Alec Radford, Gretchen Krueger, Jong~Wook Kim, Sarah Kreps, et~al.
\newblock Release strategies and the social impacts of language models.
\newblock \emph{arXiv preprint arXiv:1908.09203}, 2019.

\bibitem[Tang et~al.(2023)Tang, Chuang, and Hu]{tang2023science}
Ruixiang Tang, Yu-Neng Chuang, and Xia Hu.
\newblock The science of detecting llm-generated texts.
\newblock \emph{arXiv preprint arXiv:2303.07205}, 2023.

\bibitem[Tian(2023)]{gptzero}
E~Tian, 2023.
\newblock URL \url{https://gptzero.me}.

\bibitem[Verma et~al.(2023)Verma, Fleisig, Tomlin, and
  Klein]{verma2023ghostbuster}
Vivek Verma, Eve Fleisig, Nicholas Tomlin, and Dan Klein.
\newblock Ghostbuster: Detecting text ghostwritten by large language models.
\newblock \emph{arXiv preprint arXiv:2305.15047}, 2023.

\bibitem[Wei et~al.(2022)Wei, Wang, Schuurmans, Bosma, Xia, Chi, Le, Zhou,
  et~al.]{wei2022chain}
Jason Wei, Xuezhi Wang, Dale Schuurmans, Maarten Bosma, Fei Xia, Ed~Chi, Quoc~V
  Le, Denny Zhou, et~al.
\newblock Chain-of-thought prompting elicits reasoning in large language
  models.
\newblock \emph{Advances in Neural Information Processing Systems},
  35:\penalty0 24824--24837, 2022.

\bibitem[Zhang et~al.(2022)Zhang, Roller, Goyal, Artetxe, Chen, Chen, Dewan,
  Diab, Li, Lin, et~al.]{zhang2022opt}
Susan Zhang, Stephen Roller, Naman Goyal, Mikel Artetxe, Moya Chen, Shuohui
  Chen, Christopher Dewan, Mona Diab, Xian Li, Xi~Victoria Lin, et~al.
\newblock Opt: Open pre-trained transformer language models.
\newblock \emph{arXiv preprint arXiv:2205.01068}, 2022.

\bibitem[Zhou et~al.(2023)Zhou, Jiang, Cui, Wang, Xiao, Hou, Cotterell, and
  Sachan]{zhou2023recurrentgpt}
Wangchunshu Zhou, Yuchen~Eleanor Jiang, Peng Cui, Tiannan Wang, Zhenxin Xiao,
  Yifan Hou, Ryan Cotterell, and Mrinmaya Sachan.
\newblock Recurrentgpt: Interactive generation of (arbitrarily) long text.
\newblock \emph{arXiv preprint arXiv:2305.13304}, 2023.

\bibitem[Zhou et~al.(2022)Zhou, Muresanu, Han, Paster, Pitis, Chan, and
  Ba]{zhou2022large}
Yongchao Zhou, Andrei~Ioan Muresanu, Ziwen Han, Keiran Paster, Silviu Pitis,
  Harris Chan, and Jimmy Ba.
\newblock Large language models are human-level prompt engineers.
\newblock \emph{arXiv preprint arXiv:2211.01910}, 2022.

\bibitem[Zou et~al.(2023)Zou, Wang, Kolter, and Fredrikson]{zou2023universal}
Andy Zou, Zifan Wang, J~Zico Kolter, and Matt Fredrikson.
\newblock Universal and transferable adversarial attacks on aligned language
  models.
\newblock \emph{arXiv preprint arXiv:2307.15043}, 2023.

\end{thebibliography}
